\documentclass[times,twocolumn,final]{elsarticle}
\usepackage{framed,multirow}
\usepackage{comment}
\usepackage{amssymb}
\usepackage{latexsym}
\usepackage{subfig}
\usepackage{rotating}
\usepackage{url}
\usepackage{xcolor}
\usepackage[super]{nth}
\usepackage{float}
\usepackage{hyperref}
\usepackage{amsmath}
\usepackage{pdflscape,multicol,blindtext}
\usepackage{lipsum}
\usepackage{array,colortbl,xcolor}
\usepackage{adjustbox}
\usepackage{threeparttable}

\definecolor{newcolor}{rgb}{.8,.349,.1}

\usepackage[normalem]{ulem}

 \makeatletter
 \def\ps@pprintTitle{%
  \let\@oddhead\@empty
  \let\@evenhead\@empty
  \def\@oddfoot{\reset@font\hfil\thepage\hfil}
  \let\@evenfoot\@oddfoot
 }

\long\def\pprintMaketitle{\clearpage
  \iflongmktitle\if@twocolumn\let\columnwidth=\textwidth\fi\fi
  \resetTitleCounters
  \def\baselinestretch{1}%
  \printFirstPageNotes
  \begin{\elsarticletitlealign}%
 \thispagestyle{pprintTitle}%
    \leavevmode\vfill % ADDED BY ME
   \def\baselinestretch{1}%
    \Large\@title\par\vskip18pt%
    \ifx\@elsarticlenewpageafter\newpage@after@title% %*%
      \newpage
    \fi%
    \ifdoubleblind
      \vspace*{2pc}
    \else
      \normalsize\elsauthors\par\vskip10pt
      \footnotesize\itshape\elsaddress\par\vskip36pt
    \fi
    \ifx\@elsarticlenewpageafter\newpage@after@author% %*%
      \newpage
    \fi%
    \vfill % ADDED BY ME
    \clearpage % ADDED BY ME
    \hrule\vskip12pt
    \ifvoid\absbox\else\unvbox\absbox\par\vskip10pt\fi
    \ifvoid\keybox\else\unvbox\keybox\par\vskip10pt\fi
    \hrule\vskip12pt
    \ifx\@elsarticlenewpageafter\newpage@after@abstract% %*%
      \newpage
    \fi%
    \end{\elsarticletitlealign}%
    \gdef\thefootnote{\arabic{footnote}}%
  }

 \makeatother
\begin{document}

%\verso{Jo\~{a}o Cartucho \textit{et~al.}}

\begin{frontmatter}

\title{SurgT challenge: Benchmark of Soft-Tissue Trackers for Robotic Surgery}

%\author{Jo\~ao Cartucho$^{1}$, Alistair Weld$^{1}$, Sandro Queir\'os$^{2}$, Samyakh Tukra$^{1}$, Hiroki Matsuzaki$^{3}$, Taiyo Ishikawa$^{3}$, Minjun Kwon$^{4}$, Yongeun Jang$^{4}$, Kwang-Ju Kim$^{4}$, Gwang Lee$^{5}$, Bizhe Bai$^{6}$, Lueder Kahrs$^{6}$, Lars Boecking$^{7}$, Simeon Allmendinger$^{7}$, Leopold Mueller$^{7}$, Yitong Zhang$^{8}$, Yueming Jin$^{8}$, Bano Sophia$^{8}$, Francisco Vasconcelos$^{8}$, Leonardo Ayala$^{9}$, Wolfgang Reiter$^{10}$, Jonas Hajek$^{10}$, Stamatia Giannarou$^{1}$}

\author[1]{Jo\~{a}o Cartucho} 
\author[1]{Alistair Weld}
\author[1]{Samyakh Tukra} 
\author[1]{Haozheng Xu}
\author[2]{Hiroki Matsuzaki} 
\author[2]{Taiyo Ishikawa}
\author[3]{Minjun Kwon} 
\author[3]{Yong Eun Jang}
\author[3]{Kwang-Ju Kim}
\author[4]{Gwang Lee} 
\author[5]{Bizhe Bai} 
\author[5]{Lueder Kahrs} 
\author[6]{Lars Boecking} 
\author[6]{Simeon Allmendinger} 
\author[6]{Leopold M{\"u}ller} 
\author[7]{Yitong Zhang} 
\author[7]{Yueming Jin} 
\author[7]{Sophia Bano} 
\author[7]{Francisco Vasconcelos} 
\author[8]{Wolfgang Reiter} 
\author[8]{Jonas Hajek}
\author[9,10,11]{Bruno Silva}
%\author[12]{Lukas R. Buschle}
\author[9,10]{Estev\~{a}o Lima}
\author[11]{Jo\~{a}o L. Vila\c{c}a}
\author[9,10]{Sandro Queir\'{o}s} 
\author[1]{Stamatia Giannarou}

\address[1]{The Hamlyn Centre for Robotic Surgery, Imperial College London, United Kingdom}
\address[2]{Company Name: Jmees, Japan}
\address[3]{Company Name: Electronics and Telecommunications Research Institute (ETRI), Daejeon, South Korea}
\address[4]{Ajou University, Gyeonggi-do, South Korea}
\address[5]{Medical Computer Vision and Robotics Lab, University of Toronto, Canada}
\address[6]{Karlsruher Institut f\"{u}r Technologie: (KIT), Germany}
\address[7]{Surgical Robot Vision, University College London, United Kingdom}
\address[8]{Company Name: RIWOlink GmbH, Munich, Germany}
\address[9]{Life and Health Sciences Research Institute (ICVS), School of Medicine, University of Minho, Braga, Portugal}
\address[10]{ICVS/3B’s - PT Government Associate Laboratory, Braga/Guimarães, Portugal}
\address[11]{2Ai - School of Technology, IPCA, Barcelos, Portugal}
%\address[12]{KARL STORZ SE \& Co. KG, Tuttlingen, Germany}

%\received{--}
%\finalform{--}
%\accepted{--}
%\availableonline{--}
%\communicated{--}

\begin{abstract}
%%%
This paper introduces the ``SurgT: Surgical Tracking" challenge which was organised in conjunction with the 25\textsuperscript{th} International Conference on Medical Image Computing and Computer-Assisted Intervention (MICCAI 2022). There were two purposes for the creation of this challenge: (1) the establishment of the first standardised benchmark for the research community to assess soft-tissue trackers; and (2) to encourage the development of unsupervised deep learning methods, given the lack of annotated data in surgery. A dataset of 157 stereo endoscopic videos from 20 clinical cases, along with stereo camera calibration parameters, have been provided. Participants were assigned the task of developing algorithms to track the movement of soft tissues, represented by bounding boxes, in stereo endoscopic videos. At the end of the challenge, the developed methods were assessed on a previously hidden test subset. This assessment uses benchmarking metrics that were purposely developed for this challenge, to verify the efficacy of unsupervised deep learning algorithms in tracking soft-tissue. %\add{A key metric was the Expected Average Overlap (EAO) score, serving as an estimator of the average overlap between a tracker's bounding box and the ground truth. The EAO score, thereby, determines the overall quality of the tracker and is used to rank the competing trackers.} 
The metric used for ranking the methods was the Expected Average Overlap (EAO) score, which measures the average overlap between a tracker's and the ground truth bounding boxes.
Coming first in the challenge was the deep learning submission by ICVS-2Ai with a superior EAO score of 0.617. This method employs ARFlow to estimate unsupervised dense optical flow from cropped images, using photometric and regularization losses. Second, Jmees with an EAO of 0.583, uses deep learning for surgical tool segmentation on top of a non-deep learning baseline method: CSRT. CSRT by itself scores a similar EAO of 0.563. The results from this challenge show that currently, non-deep learning methods are still competitive. The dataset and benchmarking tool created for this challenge have been made publicly available at \url{https://surgt.grand-challenge.org/}. This challenge is expected to contribute to the development of autonomous robotic surgery and other digital surgical technologies.
\end{abstract}

%\begin{keyword}
%% MSC codes here, in the form: \MSC code \sep code
%% or \MSC[2008] code \sep code (2000 is the default)
%\MSC 41A05\sep 41A10\sep 65D05\sep 65D17
%% Keywords
%\KWD Soft-tissue Tracking\sep Unsupervised learning \sep Robotic-assisted Minimally Invasive Surgery
%\end{keyword}

\end{frontmatter}

%\linenumbers

%% main text
\section{Introduction} \label{introduction}

Tracking soft-tissue is a crucial task in Computer-Assisted Interventions (CAI), with a range of applications including autonomous tissue scanning \citep{zhan2020autonomous, wang2022towards}, autonomous tissue manipulation \citep{wang2018unified}, and other autonomous tasks in general. Specifically, for all autonomy levels at task autonomy and above, tracking is essential, as the robot operates surgical instruments in relation to the target tissue area \citep{yang2017medical}. Besides enabling autonomy, tracking soft-tissue also allows estimating tissue deformations from the endoscopic video \citep{Giannaroy2016}, which is crucial for augmented reality applications, such as the overlay of tumours or blood vessels on the endoscopic video for surgical guidance \citep{Nicolau2011AugmentedRI, Collins2020AugmentedRG}. Medical applications require accurate trackers that are robust to the dynamic conditions of surgical scenes. Therefore, prior to being utilized in real-world practice, tissue trackers need to be evaluated on a large and diverse surgical dataset that captures multiple challenges present in surgery, such as when the tracked region is occluded by a surgical instrument. In general, the problem of tracking has been extensively studied within the computer vision community, focusing mainly on natural scenes. An example of this is the Visual Object Tracking (VOT) challenge \citep{VOT_TPAMI}, which provides a benchmarking platform for tracking algorithms in non-medical contexts. While such non-medical tracking challenges exist and have driven significant progress in the field, a similar benchmark for tracking soft tissue in surgical scenes is notably absent. This lack of a surgical-specific tracking challenge has limited the development of robust and accurate tissue-tracking solutions. For other tasks such as surgical tissue 3D reconstruction, SCARED \citep{scared}, MTL \citep{Weld2022RegularisingDE} and EndoSLAM \citep{ozyoruk2021endoslam} have been created for supporting the development and evaluation of 3D reconstruction algorithms. Similarly, publicly available datasets have been created for surgical instrument segmentation \citep{allan20192017} or surgical workflow analysis \citep{bodenstedt2021heichole}, but when it comes to tracking soft-tissue, there is no publicly available dataset. Recognizing this gap, we have adapted the metrics and methodologies used in the VOT challenge to create ``SurgT: Surgical Tracking", the first challenge specifically tailored for soft-tissue tracking in surgical scenes. This is an important step towards supporting the development of tissue-tracking algorithms that can effectively handle the unique challenges posed by surgical data. This framework contains a collection of rules for tracking, metrics for analysis of performance, a data labelling tool and a curated dataset. This challenge is a sub-challenge of the Endoscopic Vision Challenge\footnote{More info at: \url{https://endovis.grand-challenge.org/}} organised in conjunction with the 25\textsuperscript{th} International Conference on Medical Image Computing and Computer Assisted Intervention (MICCAI 2022). The whole dataset, benchmarking framework, and labelling tool are publicly available. These can be found at \url{https://surgt.grand-challenge.org/}.

\section{Dataset and Annotation}
\subsection{Data} \label{data}

\begin{figure*}[t]
    \centering
    \includegraphics[width=\textwidth,height=11cm]{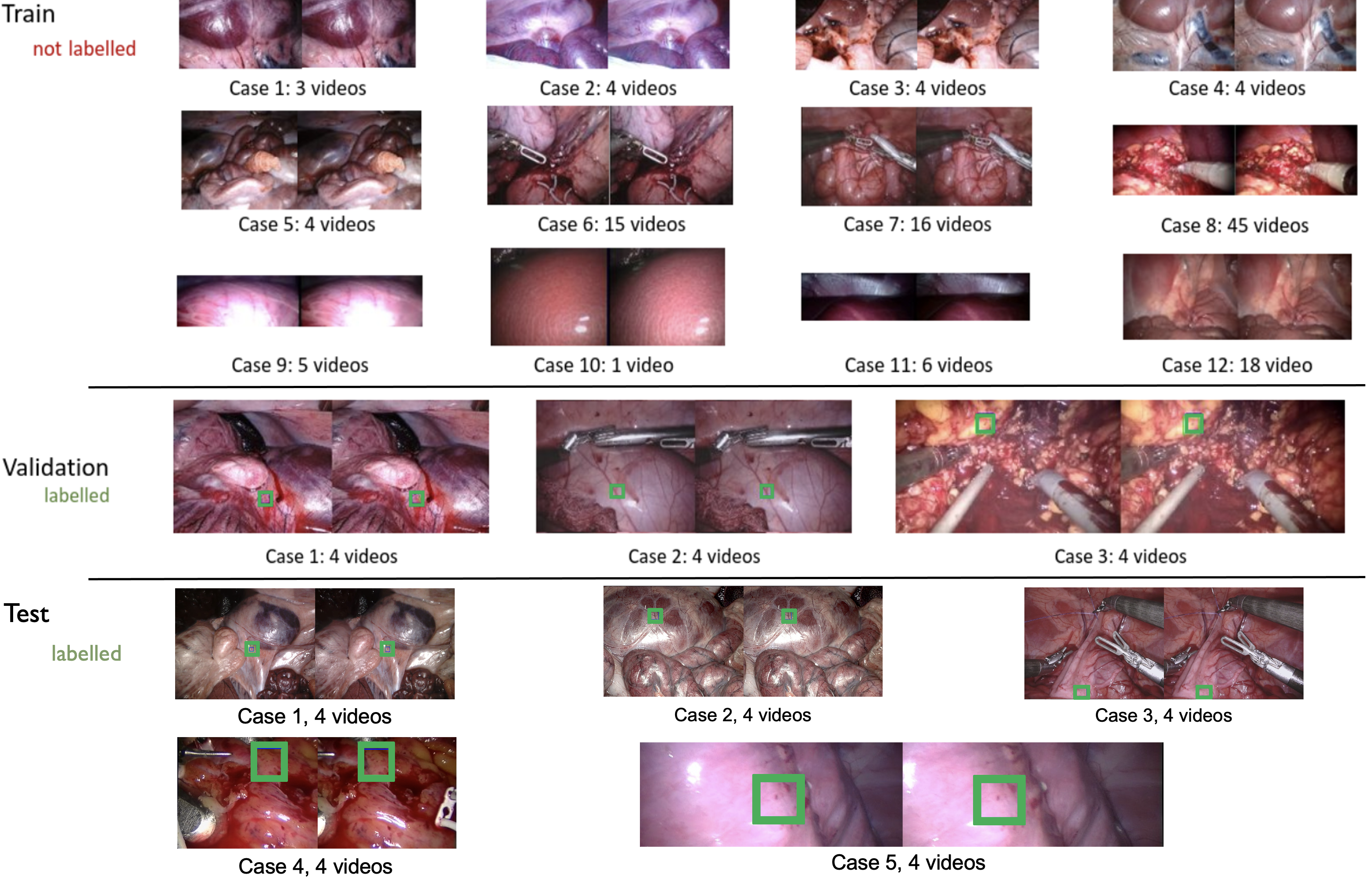}
    \caption{Sample images from all sources. Note that the training subset is not labelled, since SurgT's goal is to encourage the development of \textbf{unsupervised} algorithms. Challenging conditions in the data include occlusions due to surgical instruments, specular highlights, deformation, blood or smoke; different resolutions; different video encodings; and different illumination conditions. Note that there are no videos of phantoms, only videos from real surgeries or videos from a fresh porcine cadaver are included. Stereo camera calibration parameters and information regarding each surgery are also provided with each video.
    }
    \label{fig:data_sample}
\end{figure*}

The SurgT dataset is comprised of 157 stereo videos, with camera calibration parameters available, from 20 clinical cases. These videos were split into three subsets: (i) training - 125 videos from 12 cases; (ii) validation - 12 videos from 3 cases; and (iii) testing (hidden during the challenge) - 20 videos from 5 cases. Figure \ref{fig:data_sample} shows sample images from all cases. Every case contains multiple videos from the same surgery, and each case corresponds to a different surgery.  Across the dataset, the average duration of a video is 30 seconds, with a standard deviation of 10 seconds. The framerate of most videos is 25 $Hz$ but there are a few cases with 30 $Hz$. The video resolutions also vary. While most cases have a resolution of $1280\times1024$ $pixels$, there are videos with resolutions as low as $360\times288$ $pixels$. Finally, most videos are compressed using H.264 encoding and a few have no compression. It is important to note that while a majority of the videos in our dataset were captured by a da Vinci robot, there are also videos from the Hamlyn dataset \citep{Giannarou} that were captured using a variety of laparoscopes, including some handheld cases. As a result of this, our dataset encompasses a wide range of scenarios. This choice was intentional; our aim was to challenge the tissue tracking algorithms with a diverse set of surgical scenes, reflecting the breadth of real-world conditions and thereby enabling a comprehensive evaluation of their performance.

\begin{figure}
    \centering
    \includegraphics[width=0.65\columnwidth]{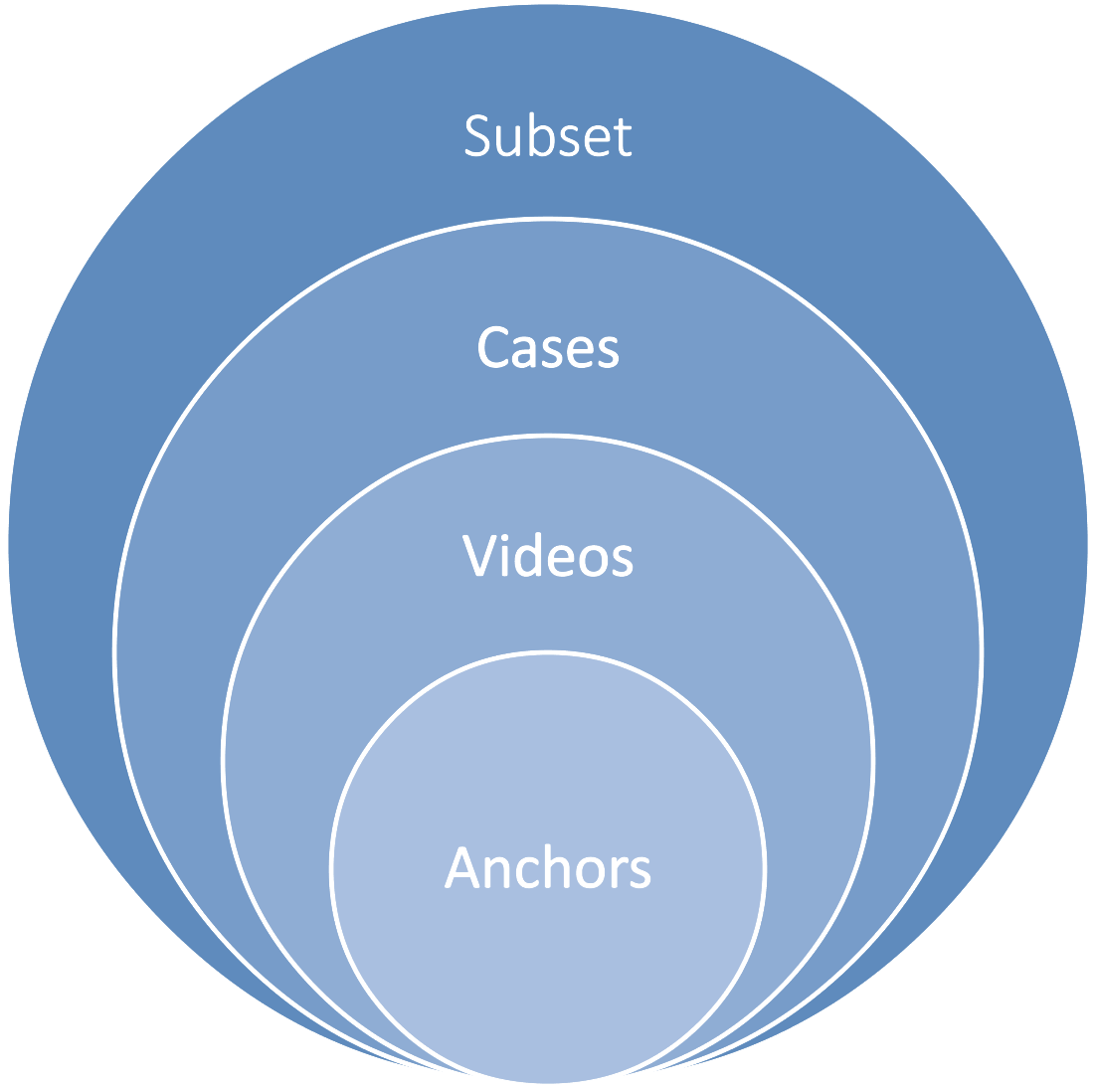}
    \caption{The SurgT dataset is comprised of training, validation and test subsets. Each subset contains all videos belonging to a number of cases. Each case contains a series of videos, where each video contains a series of anchor points (i.e. the frames where the tracker is initialized along the video, explained later in Sec. \ref{anchors}). Note that if multiple bounding boxes were labelled per video, then there would be another layer between \textit{Anchors} and \textit{Videos}. Each bounding box would have its own set of anchors. In this challenge, a single bounding box was labelled per video, hence the lack of a \textit{Bounding box} layer.
    }
    \label{fig:data_structure}
\end{figure}

The structural organization of SurgT's dataset is illustrated in Figure \ref{fig:data_structure}. For each clinical case, a calibrated stereo endoscopic camera was used to capture the videos. The stereo camera calibration parameters are available for all the videos in the dataset, which can be used to rectify the stereo-frames. The video data is not rectified by default to preserve the original videos. Code is provided for rectification.

\subsubsection{Sources of data}
The dataset was generated using the following three sources: Hamlyn dataset \citep{Giannarou}, SCARED \citep{scared}, and Kidney boundary dataset \citep{kidney_db}.

Characteristics of the validation subset are presented in Table \ref{tab:validation_dataset_description}.  On the validation subset, Case 1 is taken from SCARED \citep{scared}, which captures a fresh porcine cadaver, therefore the scene is rigid and there are no deformations or movement besides the camera's. Case 2 \citep{kidney_db} is from the Kidney boundary dataset, which captures an \textit{in vivo} porcine abdominal kidney procedure. Case 3 captures an \textit{in vivo} human surgery, concretely a Robotic-assisted Partial Nephrectomy \citep{ye2017self}.

\begin{figure}[ht]
    \centering
    \includegraphics[width=\columnwidth]{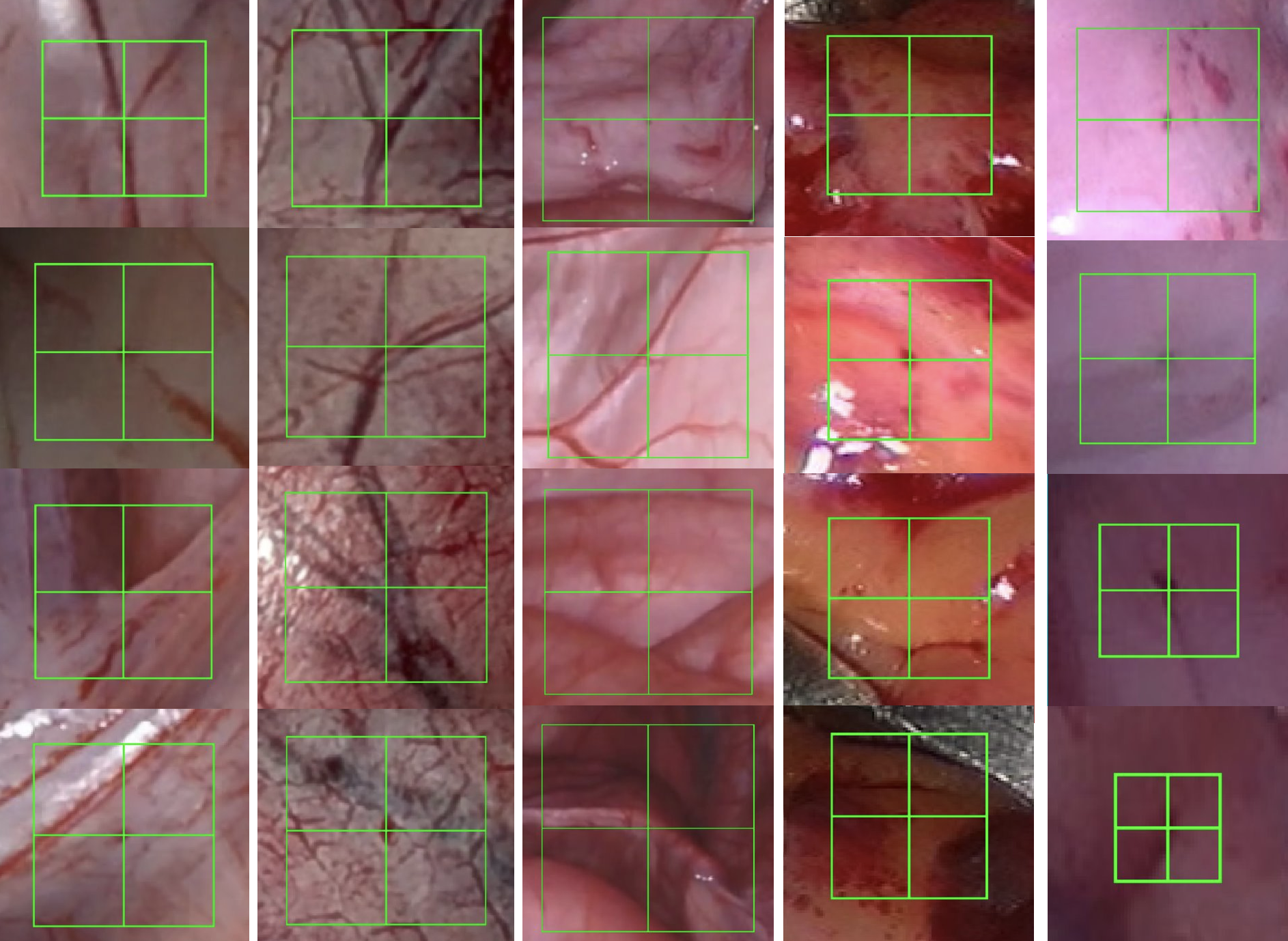}
    \caption{Bounding box samples from the hidden test subset. Each column corresponds to a case, from 1 (on the left) to 5 (on the right).
    }
    \label{fig:bbox_samples}
\end{figure}

Characteristics of the test subset are presented in Table \ref{tab:test_dataset_description}. On the test subset, the first two cases were taken from SCARED \citep{scared}. Case 3 \citep{kidney_db} from the Kidney boundary dataset. Case 4 \citep{stoyanov2005soft} is an \textit{in vivo} human Totally Endoscopic Coronary Artery Bypass (TECAB) graft. Lastly, case 5 \citep{mountney2010three} was obtained in an \textit{in vivo} porcine abdominal surgery with general camera motion and tissue-tool interaction. Figure \ref{fig:bbox_samples} shows bounding box samples from one of the anchor sequences of each case and video.

\begin{table*}[]
\caption{Description of the cases on the validation subset.}
\centering
\label{tab:validation_dataset_description}
\begin{threeparttable}
\begin{tabular}{lccc}%
  & \multicolumn{3}{c}{}  \\ \cline{2-4} 
\multicolumn{1}{l|}{} & \multicolumn{1}{c|}{Case 1} & \multicolumn{1}{c|}{Case 2} & \multicolumn{1}{c|}{Case 3} \\ \hline
\multicolumn{1}{|l|}{\textit{Data Source}}  & \multicolumn{1}{c|}{\citep{scared}} & \multicolumn{1}{c|}{\citep{kidney_db}}  & \multicolumn{1}{c|}{\citep{ye2017self}}\\ \hline
\multicolumn{1}{|l|}{\begin{tabular}[c]{@{}l@{}}\textit{Resolution}\\ \textit{W} x \textit{H} {[}$pixels${]}\end{tabular}} & \multicolumn{1}{c|}{1280 x 1024} & \multicolumn{1}{c|}{1280 x 1024} & \multicolumn{1}{c|}{480 x 270} \\ \hline
\multicolumn{1}{|l|}{\textit{Framerate [$Hz$]}} & \multicolumn{1}{c|}{25} & \multicolumn{1}{c|}{30} & \multicolumn{1}{c|}{25} \\ \hline
\multicolumn{1}{|l|}{\textit{Deforming Tissue}}   & \multicolumn{1}{c|}{-}  & \multicolumn{1}{c|}{\checkmark}      & \multicolumn{1}{c|}{\checkmark} \\ \hline
\multicolumn{1}{|l|}{\textit{Handheld Camera}}   & \multicolumn{1}{c|}{-}  & \multicolumn{1}{c|}{-}      & \multicolumn{1}{c|}{-} \\ \hline
\multicolumn{1}{|l|}{\begin{tabular}[c]{@{}l@{}}\textit{Avg. 2D velocity}\\ {[}$pixels$/$frame${]}\end{tabular}} & \multicolumn{1}{c|}{3.2} & \multicolumn{1}{c|}{2.5} & \multicolumn{1}{c|}{1.0} \\ \hline
\multicolumn{1}{|l|}{\begin{tabular}[c]{@{}l@{}}\textit{Avg. 3D velocity}\\ {[}$mm$/$frame${]}\end{tabular}} & \multicolumn{1}{c|}{0.63} & \multicolumn{1}{c|}{\cellcolor{orange!25}3.22} & \multicolumn{1}{c|}{\cellcolor{orange!25}1.94} \\ \hline
\multicolumn{1}{|l|}{\begin{tabular}[c]{@{}l@{}}\textit{Avg. Distance}\\ {Range [}$mm${]}\end{tabular}} & \multicolumn{1}{c|}{28.5} & \multicolumn{1}{c|}{\cellcolor{yellow!25}72.7} & \multicolumn{1}{c|}{27.2} \\ \hline
\multicolumn{1}{|l|}{$\%$ \textit{``ignore'' frames}} & \multicolumn{1}{c|}{1} & \multicolumn{1}{c|}{\cellcolor{green!25}28} & \multicolumn{1}{c|}{\cellcolor{green!25}33}  \\ \hline
\multicolumn{1}{|l|}{\textit{\# Discontinuities}} & \multicolumn{1}{c|}{1} & \multicolumn{1}{c|}{\cellcolor{blue!25}42} & \multicolumn{1}{c|}{\cellcolor{blue!25}22}  \\ \hline
\multicolumn{1}{|l|}{\textit{Avg. NCC score}}  & \multicolumn{1}{c|}{0.97} & \multicolumn{1}{c|}{1.0} & \multicolumn{1}{c|}{0.97} \\ \hline
\end{tabular}
\end{threeparttable}
\end{table*}

\begin{table*}[]
\caption{Description of the cases on the test subset.}
\centering
\label{tab:test_dataset_description}
\begin{threeparttable}
\begin{tabular}{lccccc}
  & \multicolumn{5}{c}{}  \\ \cline{2-6} 
\multicolumn{1}{l|}{} & \multicolumn{1}{c|}{Case 1} & \multicolumn{1}{c|}{Case 2} & \multicolumn{1}{c|}{Case 3} & \multicolumn{1}{c|}{Case 4} & \multicolumn{1}{c|}{Case 5} \\ \hline
\multicolumn{1}{|l|}{\textit{Data Source}}  & \multicolumn{2}{c|}{\citep{scared}} & \multicolumn{1}{c|}{\citep{kidney_db}}  & \multicolumn{1}{c|}{\citep{stoyanov2005soft}} & \multicolumn{1}{c|}{\citep{mountney2010three}}\\ \hline
\multicolumn{1}{|l|}{\begin{tabular}[c]{@{}l@{}}\textit{Resolution}\\ \textit{W} x \textit{H} {[}$pixels${]}\end{tabular}} & \multicolumn{1}{c|}{1280 x 1024} & \multicolumn{1}{c|}{1280 x 1024} & \multicolumn{1}{c|}{1280 x 1024} & \multicolumn{1}{c|}{360 x 288} & \multicolumn{1}{c|}{720 x 288} \\ \hline
\multicolumn{1}{|l|}{\textit{Framerate [$Hz$]}} & \multicolumn{1}{c|}{25} & \multicolumn{1}{c|}{25} & \multicolumn{1}{c|}{30} & \multicolumn{1}{c|}{25} & \multicolumn{1}{c|}{25} \\ \hline
\multicolumn{1}{|l|}{\textit{Deforming Tissue}}   & \multicolumn{1}{c|}{-}    & \multicolumn{1}{c|}{-}         & \multicolumn{1}{c|}{\checkmark}      & \multicolumn{1}{c|}{\checkmark}      & \multicolumn{1}{c|}{\checkmark}      \\  \hline
\multicolumn{1}{|l|}{\textit{Handheld Camera}}   & \multicolumn{1}{c|}{-}    & \multicolumn{1}{c|}{-}      & \multicolumn{1}{c|}{-}      & \multicolumn{1}{c|}{-}      & \multicolumn{1}{c|}{\cellcolor{red!25}\checkmark}      \\ \hline
\multicolumn{1}{|l|}{\begin{tabular}[c]{@{}l@{}}\textit{Avg. 2D velocity}\\ {[}$pixels$/$frame${]}\end{tabular}} & \multicolumn{1}{c|}{4.6} & \multicolumn{1}{c|}{6.2} & \multicolumn{1}{c|}{1.6}       & \multicolumn{1}{c|}{3.0}       & \multicolumn{1}{c|}{5.8}       \\ \hline
\multicolumn{1}{|l|}{\begin{tabular}[c]{@{}l@{}}\textit{Avg. 3D velocity}\\ {[}$mm$/$frame${]}\end{tabular}} & \multicolumn{1}{c|}{0.9} & \multicolumn{1}{c|}{0.7} & \multicolumn{1}{c|}{0.2} & \multicolumn{1}{c|}{0.5} & \multicolumn{1}{c|}{\cellcolor{orange!25}1.7} \\ \hline
\multicolumn{1}{|l|}{\begin{tabular}[c]{@{}l@{}}\textit{Avg. Distance}\\ {Range [}$mm${]}\end{tabular}} & \multicolumn{1}{c|}{29.5} & \multicolumn{1}{c|}{39.9} & \multicolumn{1}{c|}{2.9} & \multicolumn{1}{c|}{6.2} & \multicolumn{1}{c|}{28.7} \\ \hline
\multicolumn{1}{|l|}{$\%$ \textit{``ignore" frames}} & \multicolumn{1}{c|}{7} & \multicolumn{1}{c|}{\cellcolor{green!25}20} & \multicolumn{1}{c|}{5} & \multicolumn{1}{c|}{0}       & \multicolumn{1}{c|}{\cellcolor{green!25}17}       \\ \hline
%\multicolumn{1}{|l|}{$\%$ \textit{unviewable}} & \multicolumn{1}{c|}{} & \multicolumn{1}{c|}{} & \multicolumn{1}{c|}{} & \multicolumn{1}{c|}{}       & \multicolumn{1}{c|}{}       \\ \hline
\multicolumn{1}{|l|}{\textit{\# Discontinuities}} & \multicolumn{1}{c|}{3} & \multicolumn{1}{c|}{6} & \multicolumn{1}{c|}{\cellcolor{blue!25}16} & \multicolumn{1}{c|}{1} & \multicolumn{1}{c|}{\cellcolor{blue!25}17} \\ \hline
\multicolumn{1}{|l|}{\textit{Avg. NCC score}}  & \multicolumn{1}{c|}{0.97} & \multicolumn{1}{c|}{0.94} & \multicolumn{1}{c|}{0.99}       & \multicolumn{1}{c|}{0.98}       & \multicolumn{1}{c|}{0.99}       \\ \hline
\end{tabular}
\begin{tablenotes}
    \item \textit{W}: width; \textit{H}: height; \textit{Avg.}: Average; \textit{2D velocity}: represents the keypoint's velocity on the image plane; \textit{3D velocity}: represents the keypoint's velocity when projected into 3D space; \textit{Distance Range}: represents the average range of distances from the camera to the tracked keypoint; \textit{``ignore" frames}: are the frames flagged as difficult or where the bounding box is occluded/out-of-view in either the left or right image; \textit{\# Discontinuities}: represents the number of counts where the tracking algorithm could not be compared with the ground truth due to one or a sequence of ``ignore" frames;  \textit{NCC}: represents the Normalized Cross Correlation, computed using the anchor's bounding box as template, and then comparing this template with all the subsequent ground truth bounding boxes of that anchor sequence.
\end{tablenotes}
\end{threeparttable}
\end{table*}

\subsection{Data annotation protocol} \label{sec:annotation}

For the purpose of performance evaluation, the validation and test subsets were annotated. For the validation dataset, approximately 10,000 frames were annotated with stereo labels, while the test dataset had 15,000 such frames. This brings the total to 25,000 manually annotated frames. A labelling tool \citep{LABELLINGTOOL} was specifically created for this purpose. During data annotation with this tool, the objective for the annotator is to select a target keypoint (in both stereo images) and manually track and label it across the whole video. A bounding box for each keypoint is then automatically drawn around the point. The method for determining the size of the bounding box is explained in Section \ref{sec:kpt_to_bbox}.

Prior to labelling, the annotator is advised to watch the entire video to determine a suitable keypoint to track. Given the objective of the task, it is important to select suitable keypoints, bearing in mind that many factors can affect the difficulty, or even the possibility, of annotation. Once a suitable keypoint has been determined, the annotator would simply be required to select it continuously across the video, by clicking on both the left and right rectified images. The labelling tool ensures that the keypoint labelling between the stereo images obeys the epipolar line, as the tool automatically rectifies the stereo images. The labelling of the stereo images is along a horizontal line. After annotating each keypoint, the annotator is advised to check the label quality by looking back at the previous frame in the video sequence, ensuring that the same keypoint is labelled and that there was no labelling drift from the previous to the current frame. After a video has been fully annotated, the annotations are reviewed by another expert annotator, to ensure that fair and consistent annotations are provided for the challenge.

In the case where a keypoint is difficult to annotate within a frame, or causes conflicting opinions between annotators, a Boolean labelling flag $is\_difficult$ is set, declaring the keypoint annotation (and hence the frame) as difficult. Frames flagged as difficult will not affect the final ranking scores as later explained in the metric sections. Additionally, if the keypoint is occluded in either the left or the right stereo image (e.g., due to obstruction by surgical instruments), or if the keypoint moves outside of the field of view, then the annotator should change the Boolean labelling variable $is\_ visible\_in\_both\_stereo$, which is $True$ by default to declare that the keypoint is visible on both stereo images.

\subsection{Estimation of bounding box size} \label{sec:kpt_to_bbox}

\begin{figure}[ht]
    \centering
    \includegraphics[width=\columnwidth]{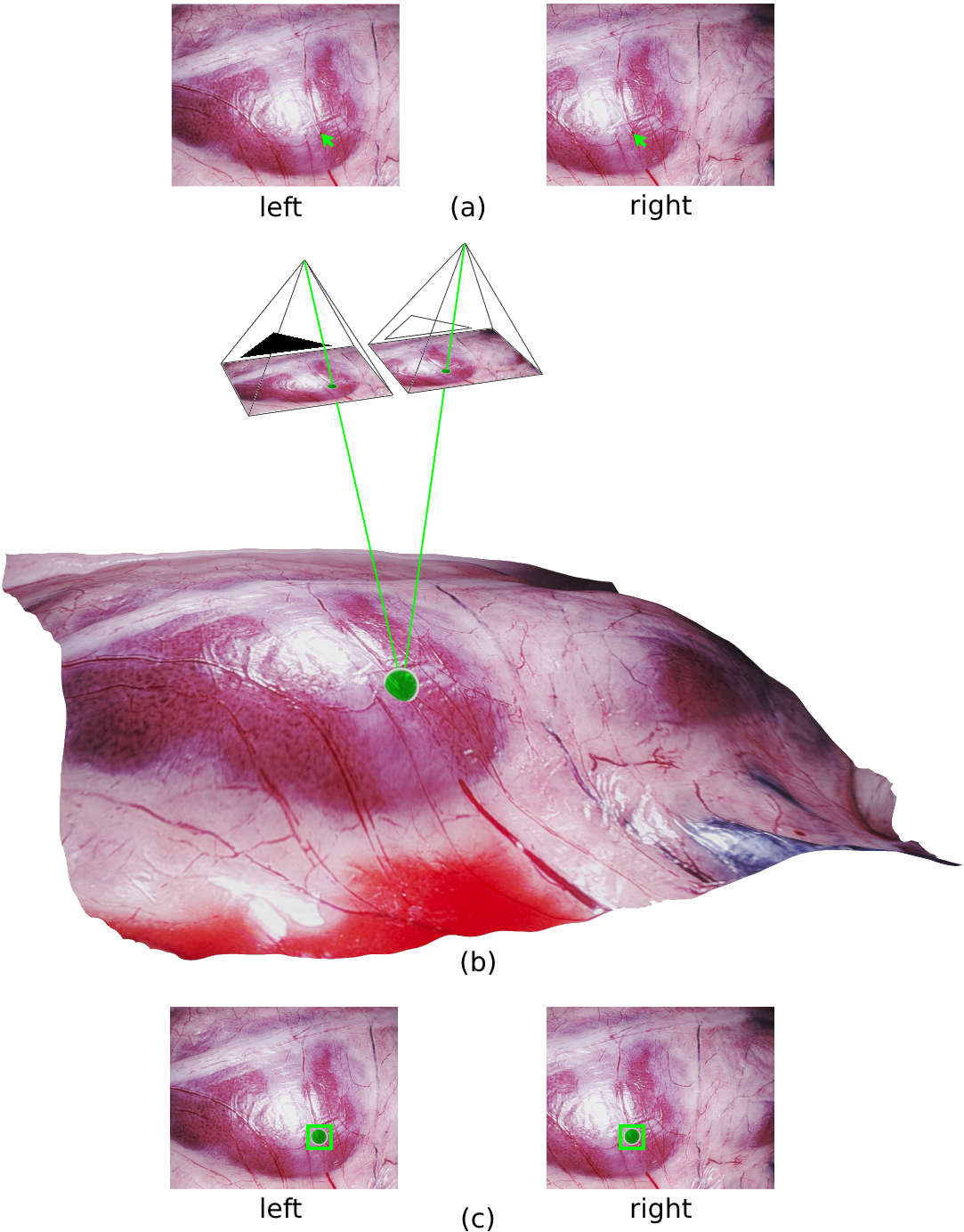}
    \caption{(a) The stereo-rectified images are annotated (green mouse cursors on top). (b) The stereo camera calibration parameters are used to project the selected keypoint into the 3D space and a virtual sphere is created around this 3D point. (c) The sphere is projected onto the 2D left/right image space and its contours are used to define the respective bounding boxes.
    }
    \label{fig:data_bbox}
\end{figure}

A bounding box encodes a target region to be tracked. Therefore, if the target moves closer to the camera, the size of the corresponding bounding box should increase. Conversely, if the target moves away from the camera, the size of the bounding box should decrease. Therefore, to increase and decrease the size of the bounding box, the disparity information from the stereo annotation was used. Concretely, the disparity is the horizontal distance between the left bounding box's centre and the right bounding box's centre, on the rectified images. Using this disparity value, the keypoint is projected into the 3D space (using Eq. \ref{ch:framework:eq:3d}) and a virtual sphere with a radius of 2.5 $mm$ is defined around this 3D keypoint. The choice of a 2.5 $mm$ radius was empirically determined, as we found this region provided sufficient information around the keypoint being tracked, while still maintaining a small area on the soft tissue - a crucial aspect for surgical applications requiring precise tracking. This sphere is then projected onto the left and right images, creating elliptical contours on each stereo image. These elliptical contours are then enclosed by a bounding box, which is used to encode the position of the target tissue region. This process is illustrated in Figure \ref{fig:data_bbox}. It should be noted that the disparity information was derived from manual ground truth annotations, and the annotators were instructed to avoid labelling drift from the previous to the current frame, thereby minimizing the potential for noise-related errors. Using this method, the size of the bounding boxes adapts to the distance between the target and the camera as desired.

\section{Benchmarking Metrics} \label{benchmarking}

The SurgT benchmarking Python toolkit is open-source and available at \url{https://github.com/Cartucho/SurgT_benchmarking}. 

\subsection{Performance evaluation protocol} 

The benchmarking metrics selected in SurgT follow the protocol to evaluate short-term trackers defined in the VOT2021 challenge \citep{kristan2021ninth}. The main reason for using the VOT metrics as a starting point was due to the maturity of the VOT2021 challenge (has been established since 2013) and for consistency.%, as it was the \nth{9} edition of this yearly challenge.

The main difference from SurgT is that VOT2021 is composed of natural videos tracking humans, cars, animals or other objects, while SurgT tracks soft-tissue surfaces. Besides that, adaptations were made to VOT's metrics, to better suit the surgical context. For example, in SurgT, the videos are not played backwards during the evaluation stage, as done in VOT2021. The reason for this is that it is unnatural to play surgical videos backwards. For example, it is unrealistic to show reverse stitching or reverse resections, or for blood to be moving inwards to ruptured blood vessels. Another difference between SurgT and VOT2021 is that the metrics were extended to support the 3D evaluation of the trackers, leveraging the availability of calibration parameters in SurgT stereo videos. In SurgT, the trackers are thus assessed, separately, in both 2D (Sec. \ref{sec:error_2d}) and 3D (Sec. \ref{sec:error_3d}). %The reason for making the stereo camera calibration parameters available is to allow the development of methods that track the soft-tissue using 3D information. 
Another crucial difference between SurgT and VOT2021 is that the former includes a ``\textit{ignore}" label to flag frames where the target region is occluded or out of the camera's field of view ($is\_ visible\_in\_both\_stereo$ $=$ $False$), or when any other visible issues are present such as motion blur ($is\_difficult$ $=$ $True$). These differences are summarized in Table~\ref{tab:diff_surgt_vot}. The following sub-sections explain in detail the performance evaluation metrics used in SurgT.

\begin{table*}[]
    \caption{Differences between SurgT and VOT2021 \citep{kristan2021ninth}.}
    \centering
    \begin{tabular}{|l|c|c|c|c|c}
      \hline
      Challenge &  \textit{Backwards} & \textit{Stereo \&} & \textit{3D}     & \textit{``Ignore''} \\
                &  \textit{video}     & \textit{cam. param} & \textit{scores} & \textit{labelling}  \\
      \hline
      VOT2021 & \checkmark & - & - & - \\
      SurgT & - & \checkmark & \checkmark  & \checkmark \\
      \hline
    \end{tabular}
    \label{tab:diff_surgt_vot}
\end{table*}

\subsection{Anchors for Tracker Initialization} \label{anchors}

\begin{figure}
    \centering
    \includegraphics[width=\columnwidth]{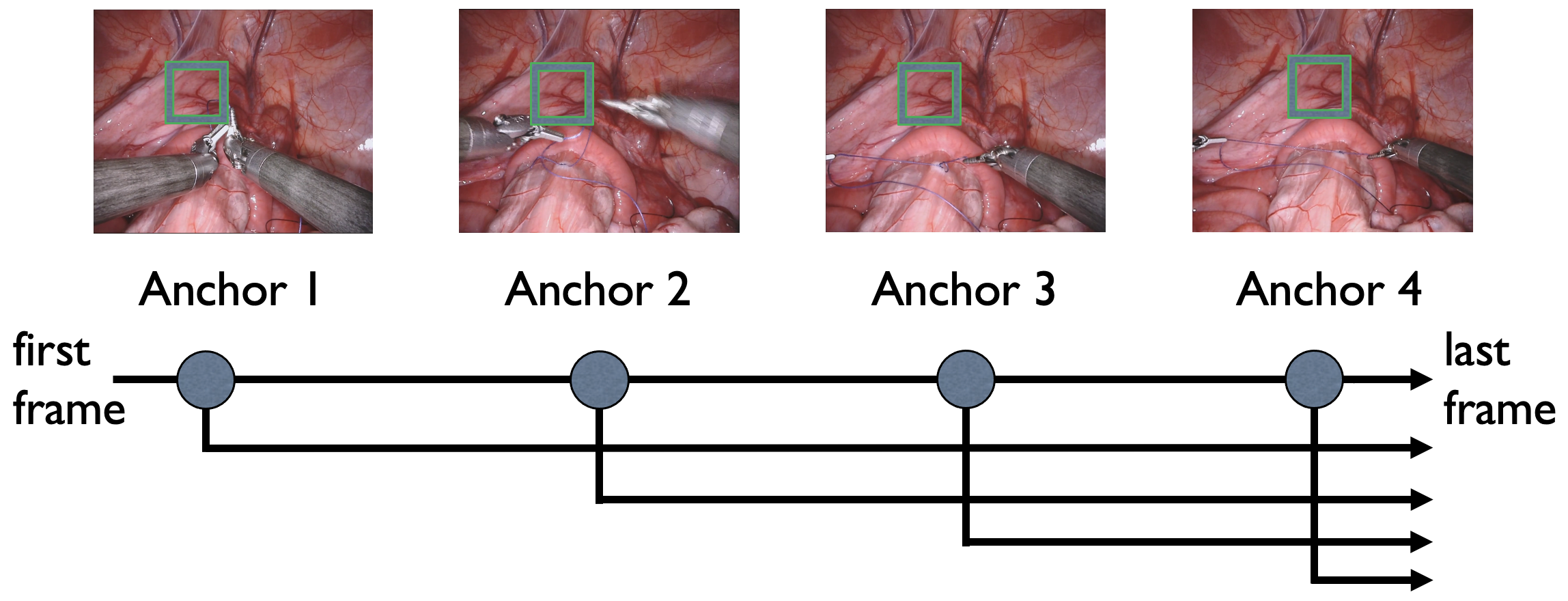}
    \caption{Anchor points used to initialize a tracker on a given video. The tracker is evaluated multiple times per video, each time initialized at a distinct anchor point and evaluated until the end of the video.
    }
    \label{fig:anchor_points}
\end{figure}

The trackers are initialized at pre-defined anchor points, which are the target regions, encoded as left/right bounding boxes, that are provided as input to a tracking algorithm. The tracker's task is to then output the target bounding box for the following sequential frames in the video, until the end of the video. After completing the video, the tracker is re-initialized using the same video, but this time starting at the next anchor point, as illustrated in Figure \ref{fig:anchor_points}. Evaluation of a video is only finalized once the tracker has been initialized and assessed using all anchors of the video. Successive anchors were roughly spaced by 50 frames (as suggested by VOT \citep{kristan2021ninth}), while always ensuring that anchor points are set on frames with clearly visible keypoints. In the test subset, there were, on average, 13 anchor points per video.

There are two main reasons for using anchor points and for spacing them in time: (i) it ensures that only suitable initial tracking targets are given, i.e. the target point is clearly visible (without blur or compression artefacts); %, instead of starting at a fixed predefined frame (e.g. the first) for all videos; 
(ii) it allows a more complete and reliable assessment of the tracker on each video, reducing the influence of individual frames in the evaluation and emphasising the video sequence as a whole. For example, if there is a tracking failure when initialized in the first anchor, it still has a chance of improving its score when initialized in the following anchor points.

\subsection{Monocular metrics}
\label{sec:error_2d}

The tracker's task is to locate the target point, encoded in a bounding box, for all video frames following the anchor point initialization. If the target point is not visible in the image, then the tracker should identify this and give a classification output \textit{None}. A target point is not visible when, for example, the target is being occluded by a surgical instrument, or when the target is outside the camera's field of view.

A frame is considered valid if the target point is both (i) visible and (ii) not difficult (see Sec. \ref{sec:annotation}). Given the valid frames, the tracker's performance is assessed by comparing the tracker's bounding box prediction with the ground truth bounding box. The tracker is assessed for all subsequent valid frames until a tracking failure is detected (if it is ever triggered), or up to the video's last valid frame.

\subsubsection{2D Tracking failure}
\label{sec:2d_track_fail}

For each anchor point, a 2D tracking failure is detected when the Intersection over Union (IoU) between the ground truth and predicted bounding box is smaller than 0.1, for 10 consecutive frames. The target point is required to be tracked in both left and right stereo images. Therefore, 2D tracking failure can be triggered if the failure occurs in either the left or the right image. After the tracking failure, the 2D accuracy and 2D error scores are not updated further for that anchor sub-sequence. Conversely, the 2D robustness score is negatively affected by tracking failure since all subsequent frames are recorded as failed to be tracked.

\subsubsection{Accuracy}

For each anchor sub-sequence, the accuracy is the average IoU between the predicted ($P$) and ground truth ($G$) bounding boxes, before any tracking failure occurs:

\begin{equation}
    IoU = \frac{|P \cap G|}{|P \cup G|}
\end{equation}

\begin{equation}
    Accuracy = \frac{1}{n} \sum_{i=1}^{n} IoU_{i}
\end{equation}

\noindent where $i$ is the frame index and $n$ is the number of frames for which tracking has been successful and ground truth is available. For example, if there are a total of 1000 frames, and the 2D tracking failure is triggered in the last frame, it means that during the last 10 frames, the IoU score was $<$ 0.1. Therefore, the accuracy will be calculated using the first 990 frames, which are the $n$ frames before tracking failure occurs. Note that $n$ only includes frames for which ground truth is available. Some frames, as explained in the annotation protocol (Sec. \ref{sec:annotation}), were flagged as difficult ($is\_difficult$ $=$ $True$) or as not visible ($is\_ visible\_in\_both\_stereo$ $=$ $False$), and thus these frames are ignored to ensure that they do not affect the accuracy score. $n$ also does not include frames where the tracker made no prediction, since failure to predict an updated bounding box should not affect its accuracy score, only its robustness score.

\subsubsection{2D Error}

For each anchor sub-sequence, the 2D error is the average distance (in $pixels$) between the centre of the predicted bounding box ($P$) and the centre of the ground truth bounding box ($G$), before tracking failure occurs:

\begin{equation}
    Error_{2D} = \frac{1}{n} \sum_{i=1}^{n} |P_{i_{2D}} - G_{i_{2D}}|
\end{equation}

\noindent where $n$ is the number of frames for which ground truth is available and tracking successful. The $n$ in the Accuracy score and $Error_{2D}$ score equations are the same.

\subsubsection{2D Robustness}

For each anchor sub-sequence, the 2D robustness score is given by the following ratio:

\begin{equation}
    Robust_{2D} = \frac{N_{successful_{2D}}}{N_{valid} + N_{excess}}
\end{equation}

\noindent where, $N_{successful_{2D}}$ is the number of tracked frames in which the IoU score is larger than 0.1, for both left and right bounding boxes; $N_{valid}$ is the number of frames where the target point was visible in both stereo images and not marked as difficult; and $N_{excess}$ is the number of frames where the tracker made a prediction but the target point was not visible in that frame. For example, if the tracker predicts a bounding box when the target point is outside the camera's field of view, the number of excessive frames is incremented. Similarly, if the tracker predicts a bounding box when the target point is occluded by a surgical instrument, the number of excessive frames is also incremented.

\subsection{Stereo metrics}
\label{sec:error_3d}
The 3D metrics are computed separately from the 2D ones, for each anchor point sequence. 

\subsubsection{Calculating the target point's 3D coordinates}

Given that the input images are rectified, and the target point is being tracked in both left and right images, it is possible to project the target point into 3D. Specifically, given the central pixel position of both left and right bounding boxes, it is first possible to calculate the stereo disparity ($d$) using:

\begin{equation}
    d = u_{left} - u_{right}
\end{equation}

\noindent where $u$ is the pixel coordinate along the epipolar line. Then, the disparity ($d$) and the left central pixel coordinate ($u_{left}$, $v_{left}$) can be used to project the point into 3D space, using the stereo calibration parameters as follows:

\begin{equation}
    s
	\begin{bmatrix} 
	X\\
	Y\\
	Z\\
	1\\
	\end{bmatrix}
    =
    \begin{bmatrix} 
	1 & 0 & 0 & -{c_x}\\
	0 & 1 & 0 & -{c_y}\\
	0 & 0 & 0 & f\\
	0 & 0 & -\frac{1}{b} & 0\\
	\end{bmatrix}
	\begin{bmatrix} 
	u_{left}\\
	v_{left}\\
	d\\
	1\\
	\end{bmatrix}
	\label{ch:framework:eq:3d}
\end{equation}

\noindent where $c_x$ and $c_y$ represent the optical centre, $f$ the focal length, $b$ the baseline of the rectified cameras, and $s$ an arbitrary scale factor. This equation is applied twice, calculating the predicted and ground truth 3D points, respectively, from the predicted and ground truth bounding boxes.

\subsubsection{3D Tracking Failure}

As previously mentioned, 2D and 3D metrics are calculated separately. Therefore, 3D tracking failure detection is also separate from 2D tracking failure. For each anchor sub-sequence, a 3D tracking failure is detected when the distance between the predicted and ground truth 3D target positions (3D error) is larger than 10 $cm$ or when the disparity is $\leq$ 0 for ten consecutive frames. Both conditions can be used together; for example, the tracker may have a 3D error $>$ 10 $cm$ for five frames, and then get a negative disparity for the next five frames, which would still trigger a 3D tracking failure.

\subsubsection{3D Error}

For each anchor sub-sequence, the 3D error is the average distance in $mm$ between the predicted ($P$) and the ground truth ($G$) bounding boxes, before 3D tracking failure occurs:

\begin{equation}
    Error_{3D} = \frac{1}{n} \sum_{i=1}^{n} |P_{i_{3D}} - G_{i_{3D}}|
\end{equation}

\noindent where, $n$ is the number of frames before 3D tracking failure on which the disparity is $\geq$ 0. Note that, similarly to the 2D metric, the set $n$ does not include frames that are marked as difficult ($is\_difficult$ $=$ $True$) or in which the target is not visible ($is\_ visible\_in\_both\_stereo$ $=$ $False$).

\subsubsection{3D Robustness}

For each anchor sub-sequence, the 3D robustness score is given by the following ratio:

\begin{equation}
    Robust_{3D} = \frac{N_{successful_{3D}}}{N_{valid} + N_{excess}}
\end{equation}

\noindent where, $N_{successful_{3D}}$ is the number of tracked frames in which the 3D error is smaller or equal to 10 $cm$, and the disparity is positive. The other two variables, $N_{valid}$ and $N_{excess}$, are the same as the ones used for the 2D robustness score.

\subsection{Weighted Average of the Results}

All final scores for each subset, case, or video on the aforementioned 2D and 3D metrics are obtained using weighted averages. For example, the 2D accuracy of a video, which contains multiple anchors, is obtained using:

\begin{equation}
    Video_{Accuracy} = \sum_{i=1}^{k} \frac{Accuracy_i \times n_i}{n_{video}}
\end{equation}

\noindent where, $k$ is the number of anchors in that video, $Accuracy_i$ is the accuracy of anchor sub-sequence $i$, $n_i$ is the number of frames in sub-sequence $i$ before 2D tracking failure is detected, and $n_{video} = \sum_{i=1}^{k} n_i$. 

For a given case, the accuracy is mathematically defined as:

\begin{equation}
    Case_{Accuracy} = \sum_{i=1}^{m} \frac{Video_{{Accuracy}_i} \times n_{{video}_i}}{n_{case}}
\end{equation}

\noindent where, $m$ is the number of videos per case and $n_{case} = \sum_{i=1}^{m} n_{{video}_i}$.

Finally, for the validation or test subsets, the accuracy is estimated as:

\begin{equation}
    Subset_{Accuracy} = \sum_{i=1}^{l} \frac{Case_{{Accuracy}_i} \times n_{{case}_i}}{n_{subset}}
\end{equation}

\noindent where, $l$ is the number of cases per subset and $n_{subset} = \sum_{i=1}^{m} n_{{case}_i}$.

\subsection{Expected Average Overlap (EAO)} \label{eao}

Although the aforementioned 2D and 3D metrics are used to assess different characteristics of the tracker, ultimately it is the Expected Average Overlap (EAO) metric that determines the overall quality of the tracker and is therefore used to rank the different trackers. The EAO is computed by averaging the IoU score from the merged sequence ($IoU_{subset}$) between $N_{min}$ and $N_{max}$, excluding all $``ignore"$ frames:
\begin{equation}
    EAO = \frac{1}{N_{max} - N_{min}} \sum_{i=N_{min}}^{N_{max}} IoU_{{subset}_i}
\end{equation}
\noindent where, $N_{min}$ and $N_{max}$ are defined as the average length of all video sequences $(IoU_{video_{i}})$ $\pm$ one standard deviation of these lengths. The same $N_{min}$ and $N_{max}$ values were applied to all trackers during evaluation. $N_{min}$ and $N_{max}$ for the test subset highlight the EAO scoring region in Fig.~\ref{fig:eao_curve}. $EAO \in [0,1]$, where a result of $1$ is the perfect score. To obtain $IoU_{subset}$ the score of all anchor sub-sequences are merged for each video, and then the scores of all videos of a subset are also merged into $IoU_{subset}$.

\subsubsection{Anchor IoU}

Each anchor point, defined in Sec. \ref{anchors}, creates a  sub-sequence of the full video. In each sub-sequence, the IoU score is calculated for each frame. For a demonstrative example, we define a video containing two anchor points and the following IoU scores for the two sub-sequences:
\begin{equation*}
  IoU_{s1} = [1, 0.8, 0.6, 0.5, ``ignore", ..., 0], len = 100
\end{equation*}
\begin{equation*}
  IoU_{s2} = [1, 1, 1, ``ignore", ``ignore", ..., 0], len = 50
\end{equation*}

As illustrated in this example, usually the IoU scores start with larger values (close to 1), since the tracker has just been initialized, and decrease as the video sequence progresses due to tracking drift (getting as low as 0). The scores of image frames that are flagged as difficult or in which the target is not visible are recorded as $``ignore"$, since these frames should not affect the final score.

Although not exemplified in the previous sub-sequences, if a 2D tracking failure is detected, then all subsequent frames of a given sub-sequence are set to 0, since the tracker already failed. This means that if the tracker drifts and 2D tracking failure is triggered, then the EAO score gets negatively impacted due to the trailing zeros. 

\subsubsection{Video IoU}

After all anchor sub-sequences have been processed and the results recorded, an average is computed to obtain an IoU sequence for the video. In this simplified example, we would obtain (given $IoU_{s1}$ and $IoU_{s2}$):
\begin{equation*}
  IoU_{video_{1}} = [1, 0.9, 0.8, 0.5, ``ignore", ..., 0], len = 100
\end{equation*}

Note that, since in the \nth{4} frame $IoU_{s1} = 0.5$ and $IoU_{s2} = ``ignore"$ , then the \nth{4} frame of $IoU_{video_{1}} = 0.5$. If all the scores of a frame are recorded as $``ignore"$ (like in the \nth{5} frame), then the respective frame in $IoU_{video_{1}}$ is also set to $``ignore"$. Also note that when merging scores, the length of the largest sub-sequence is kept to merge all the scores from the anchors of that video (in this case, $len = 100$).
%An example of such a sub-sequence would be:
%\begin{equation*}
%  IoU_{s3} = [1, 0, 0, 0, 0, 0, 0, 0, 0, 0, 0, \textbf{0}, \textbf{0}, ..., \textbf{0}], len = 150
%\end{equation*}

%In $IoU_{s3}$, there were 10 consecutive frames scoring less than the $IoU$ threshold of 0.1, triggering the 2D tracking failure and setting all remaining frames to 0 (in bold).

Let us now assume a second video, either from the same clinical case or from another case, with the following scores:
\begin{equation*}
  IoU_{video_{2}} = [1, 1, 1, 0, 0, ..., 0], len = 200
\end{equation*}

Using all recorded IoU scores, a single merged sequence ($IoU_{dataset}$) score can be calculated using all videos (here, $IoU_{video_{1}}$ and $IoU_{video_{2}}$). Note that each video contributes exactly with a single sequence of IoU scores, independently of the number of anchors in that video. This is to avoid a scenario where a video with more anchors has a bigger contribution towards the final EAO score when compared to a video with fewer anchors. In this simplified example, the subset score would be:
\begin{equation*}
  IoU_{subset} = [1, 0.95, 0.9, 0.25, 0, ..., 0], len = 200
\end{equation*}

Note that, again, the $``ignore"$ frame of $video_1$ was ignored (its \nth{5} element). %Concretely, the \nth{5} of $IoU_{video_{1}} = ``ignore"$ and the \nth{5} of $IoU_{video_{2}} = 0$, hence the \nth{5} of $IoU_{subset} = 0$.
If both videos had $``ignore"$ in the \nth{5} frame, then $``ignore"$ would still show up on the \nth{5} frame of $IoU_{subset}$, but this would not affect the final EAO score as explained later. Also note that, the length of $IoU_{subset}$ is as large as the largest video sequence (in this case $len(IoU_{video_{2}}) = 200$). This implies that larger videos, with more frames, may have a larger contribution to the final EAO score than smaller videos. However, this effect is reduced given that the final EAO score is computed over a pre-defined range of frames. 

\subsection{Winner identification protocol}

The teams were ranked according to their EAO scores on the test subset as shown in Fig.~\ref{fig:rank_graph}.

\section{Baseline methods for comparison}

Two baseline bounding box trackers were used to compare the results obtained by the participants namely, the CSRT (Sec. \ref{baseline:trad_cv}) - a traditional computer vision method, and the TransT (Sec. \ref{baseline:deep_learning}) - a deep learning method trained only on natural scenes.

\subsection{[\textbf{CSRT}] Baseline - traditional computer vision}
\label{baseline:trad_cv}

OpenCV's Channel and Spatial Reliability Tracker (CSRT) \citep{lukezic2017discriminative} is provided on the SurgT benchmarking code by default. The idea is for the participating teams to use it as a baseline for comparison, and to monitor the performance of their own algorithms during development. CSRT is a Discriminative Correlation Filters (DCF) tracker that adapts the filter response to the target region that is suitable for tracking (spatial reliability), in addition to using all the colour channels for localizing the bounding box in the next frame (channel reliability) \citep{lukezic2017discriminative}. %The `CSRT' tracker shows an impressive performance on SurgT data, hence, it is used as baseline.

\subsection{[\textbf{TransT}] Baseline - natural scenes deep learning}
\label{baseline:deep_learning}

Transformer Tracking (TransT) \citep{chen2021transformer} is a state-of-the-art tracker using an attention-based feature fusion approach, inspired by \citet{vaswani2017attention}. To be used as a baseline tracker, TransT was trained on the VOT2020 dataset \citep{kristan2020eighth} until convergence, using ground truth bounding boxes of natural scenes such as moving objects, humans, animals, and others. No surgical data was used during the training of TransT. This is used to assess whether a supervised deep learning method, trained on an irrelevant dataset, can generalise well onto SurgT.

\section{Challenge submissions} \label{submissions}

%{\color{red} CONTINUE FROM HERE}

\subsection{[\textbf{Jmees}] Company name: Jmees, Japan}

Jmees's tracking method is composed of a (a) \textbf{tracking} part and a (b) \textbf{correction} part. For (a) \textbf{tracking}, the CSRT tracker \citep{Luke_i__2018} was used along with template matching using the initial anchor's bounding box as a template. For (b) \textbf{correction}, three types of correction were applied. First, the stereo disparity is calculated from the rectified images, between the centre of the left/right bounding boxes. The ratio between the disparity of the initial frame and that of the updated  frame is  used  to  scale  the  size  of  the  updated  bounding  boxes. Second, surgical instrument segmentation is used to  judge  the occlusion. If an occlusion is detected then tracking failure is determined. The segmentation was trained using a supervised learning method and a publicly available dataset for surgical instrument segmentation \citep{maier2014can}. Third, back-tracking and template matching between current and previous frames are used for the validity assessment of the tracking. When the result of the assessment is not valid, then tracking failure is flagged.

\subsection{[\textbf{ETRI}] Company name: Electronics and Telecommunications Research Institute, Republic of Korea}

ETRI's tracker is based on the Unsupervised Deep Tracking (UDT) \cite{wang2021unsupervised}, which is built on a discriminant correlation filter. UDT is an unsupervised learning method, trained using crops of the upper central image regions. This method learns to track by comparing adjacent frames with forward and backward predictions. In ETRI's method, the template is cropped (input image to a neural network) and search patches from two consecutive frames conducting forward tracking and backward verification - which is why there is no need for ground truth, as the patches are predicted on the raw images, in adjacent frames. The upper centre part of the image was used as the target because it was assumed that there should be minimal movement and therefore it is unlikely that the target would move out of the pre-defined cropped region, in a short period. Additionally, the upper centre was used, instead of the centre, to avoid issues with surgical tools and occlusions. A dataset was created using these cropped images and split into a training and validation subset in a 90:10 ratio. The difference between the initial bounding box and the predicted bounding box formulates a consistency loss for network learning. In ETRI's method, the comparison was performed within 1 to 4 frames - choosing one to four patches as the template and the remaining as search patches. The obtained weights were learned from scratch on SurgT's training subset, without the use of any pre-trained weights.

\subsection{[\textbf{MEDCVR}] Medical Computer Vision and Robotics lab, University of Toronto, Canada} 

MEDCVR's method is based on TransT \citep{chen2021transformer}, a Siamese-based tracker. TransT solves the problem that correlation filters in Siamese-based trackers tend to lose semantic information and fall into local optima through an attention-based feature fusion network. The default TransT was trained until convergence on the VOT2020 dataset \citep{kristan2020eighth}, which does not contain any surgical data. Then, unsupervised training on the feature extraction layers was used to tune TransT to surgical data and improve accuracy, using SurgT's training subset. Inspired by previous works which do self-supervision by predicting rotation angles and context reconstruction \citep{he2022masked}, two proxy tasks were exploited: masked volume inpainting and image rotation. The masked volume inpainting is motivated by prior work which focused on 2D images \citep{pathak2016context}. The rotation prediction task predicts the angle categories by which the input sub-volume is rotated.

\subsection{[\textbf{SRV}] Surgical Robot Vision research group, University College London, United Kingdom}

SRV tackled this problem as a regression machine learning task, with images plus an initial bounding box as input and bounding box movement as output. Concretely, a receptive field twice as big as the initial bounding box is created. This receptive field defines the image region to where the bounding box can move from one frame to the next. After predicting this movement, a new receptive field is created for every two subsequent frames. The purpose of SRV's network is to estimate the bounding box's displacement. As an unsupervised task, spatial features from the input must be extracted without ground truth. A masked auto-encoder (MAE) \citep{he2022masked}, with pre-trained weights, was applied as a feature encoder. The pre-trained weights achieved a good reconstruction on the SurgT training subset, which indicates its good representation capability on  this  data and, hence, does not require further fine-tuning. Also, a ‘goal’ image \citep{andrychowicz2017hindsight} was introduced to represent the region of interest and guides the network to discover spatial differences. This ‘goal’ is also encoded by the same MAE encoder and concatenated with the bounding box feature. The two MAE concatenated features are then provided as input to a regression model (three layers of a vision transformer block \citep{dosovitskiy2020image} plus two fully connected layers) to predict the updated location of the bounding box. This model is trained by augmenting randomly sampled bounding boxes and frames from the SurgT training subset. For this augmentation, a random 2D translation displacement is applied to the sampled images. These displacements are known and can be used as the ground-truth signal for training the regression model. The network takes the augmented images and compares them with the ‘goal’ image to predict the displacements which can move the bounding box back to the original image. During inference, the ‘goal’ image is initialised with the receptive field around the target bounding box in the first frame. The extracted feature vector from the ‘goal’ receptive field is updated with every new frame using the following momentum equation: $Goal_{k+1} = mGoal_k + (1 - m) newGoal$, where, \textit{newGoal} is the feature extracted from the receptive field around the current bounding box estimation and \textit{m} is a tenable momentum parameter. The similarity between $mGoal_k$ and $newGoal$ is measured with mean square error. When $similarity<0.4$, representing occlusion or poor tracking, $m$=1 otherwise, $m$=0.95. On low-resolution videos, the tracking trajectory can suffer from jitters. Therefore, a Kalman filter designed for 2D tracking \citep{patel2013moving} is used to refine the network's regression result.

\subsection{[\textbf{KIT}] Karlsruhe Institute of Technology, Germany}

This team proposed a CNN feature map tracker, consisting of three main components: feature map generation, similarity approximation and distance measurement. First, a feature map for a specific target region in the image is generated based on a ResNet50 backbone, which was pre-trained on the ImageNet dataset \citep{deng2009imagenet}. Second, similar regions in subsequent frames are searched, creating a set of potential bounding box candidates. These are derived by exploring the surroundings of the bounding box given a predefined step-width. Third, the candidate feature maps are quantitatively compared with the target feature map based on the cosine distance metric to find the most suitable one.

\section{Post-challenge submissions}

\subsection{[\textbf{ICVS-2Ai}] Collaboration between Life and Health Sciences Research Institute (ICVS) and Applied Artificial Intelligence Laboratory (2Ai), Portugal}
This method is based on ARFlow, an unsupervised dense optical flow estimator introduced in \citet{liu2020learning}. ARFlow builds on the PWC-net method \citep{sun2018pwc}, incorporating heavy augmentation as a self-supervised regularization technique. To train this model, pairs of consecutive images from the SurgT train subset were randomly cropped to a square of size equal to the smallest dimension and resized to 256$\times$256 $pixels$. Since no ground truth data is available, the model was trained using two losses: a photometric loss applied on non-occluded pixels (estimated using a forward-backward checking method) of the reconstructed image of the current frame (obtained by wrapping the subsequent frame with the predicted optical flow); and a regularization loss term to smooth the predicted optical flow and reduce ambiguity in textureless patterns \citep{liu2020learning}. The training was performed with the AdamW optimizer over 40 epochs, using a constant learning rate schedule. The epoch with the best validation EAO score was then selected. During inference, this model was used to track the centroid of the target bounding box. To do so, the current and subsequent frames are center-cropped on the current centroid point (with size 256$\times$256 without resizing) and fed to the model. The displacement at the location of the centroid is used to update the bounding box position. In its turn, the disparity between left-right images is estimated and used to update the bounding box size (keeping the proportion calculated at the initial frame of the video). This process is consecutively repeated for every pair of frames of a given video.

\subsection{[\textbf{RIWOlink}] Company name: RIWOlink GmbH, Germany}
RIWOlink's tracking method is based on the UDT tracker \citep{Wang_2019_Unsupervised}. A siamese convolutional neural network is used to extract features from a template patch and two search patches. Tracking is implemented with Discriminative Correlation Filters regressing the search features to a Gaussian response centred at the bounding box centre. Forward and subsequent backward tracking using a pseudo label provides an unsupervised consistency loss.
The unsupervised training of the model is executed on the SurgT unlabeled training data. Left and right sequences are treated as different training samples. Adding cholec80 data~\citep{twinanda_endonet_2017} to the unsupervised training improved the validation EAO metric by 0.05. Adding even more unlabeled data lead to a decrease in the validation metrics.
The model is trained for 100 epochs with a batch size of 128, a learning rate of $1\times10^{-3}$ and a weight decay of $5\times10^{-5}$. Images are pre-processed as image patches of size $125\times125$ with mean of training split subtracted. Forward-backward error~\citep{5596017} is used to detect tracking failures.
The best model is chosen with the EAO metric evaluated on the SurgT validation subset.

\section{Challenge results} \label{results}

The detailed results on each of the test cases are presented in Table~\ref{tab:big_results}, according to the benchmarking metrics defined in Sec. \ref{benchmarking}. This table is divided into two sub-tables, one for 2D and another for 3D scores. All test scores, on all cases, are then averaged into Table~\ref{tab:test_res}. For the 2D test case scores, we have Robustness (Rob.) and Accuracy (Acc.) along with 2D Error (in $pixels$):
\begin{itemize}
    \item ICVS-2Ai: Outperformed the other methods, displaying the highest robustness score of 0.901. It maintained an accuracy of around 0.818, with an average 2D error of 6.7 $pixels$;
    \item Jmees: Consistently performed across the test cases, averaging robustness and accuracy values of 0.868 and 0.818 respectively. The method reported an average 2D error of about 5.3 $pixels$;
    \item CSRT: Presented comparable results to Jmees, with robustness and accuracy averages of 0.872 and 0.769. The error measure stands at approximately 5.7 pixels.
    \item RIWOlink, ETRI, MEDCVR, SRV, TransT and KIT: All these methods showed a relatively lower performance, particularly noticeable in their error values.
\end{itemize}

In terms of 3D test case scores, which evaluates the Robustness in terms of 3D Error (in $mm$):
\begin{itemize}
    \item ICVS-2Ai and Jmees: Both exhibited consistent results across all test cases. ICVS-2Ai marginally outperformed with a 3D error average of 2.3 $mm$, whereas Jmees trailed closely at 2.7 $mm$.
    \item CSRT and RIWOlink: Demonstrated relatively consistent results, with CSRT having a 3D error of 3.3 $mm$ and RIWOlink experiencing a slightly higher error at 5.8 $mm$.
    \item ETRI: Maintained robustness close to that of RIWOlink but had a 3D error slightly lower than RIWOlink, averaging at 5.7 $mm$.
    \item MEDCVR and SRV: Both faced challenges, with SRV recording a 3D error average of 8.5 $mm$, and MEDCVR reaching 9.2 $mm$.
    \item TransT and KIT: Were characterized by higher error values in the 3D domain, with TransT and KIT averaging at 17.3 $mm$ and 11.9 $mm$, respectively.
\end{itemize}

\begin{figure*}[h]
\centering
\includegraphics[width=1.7\columnwidth]{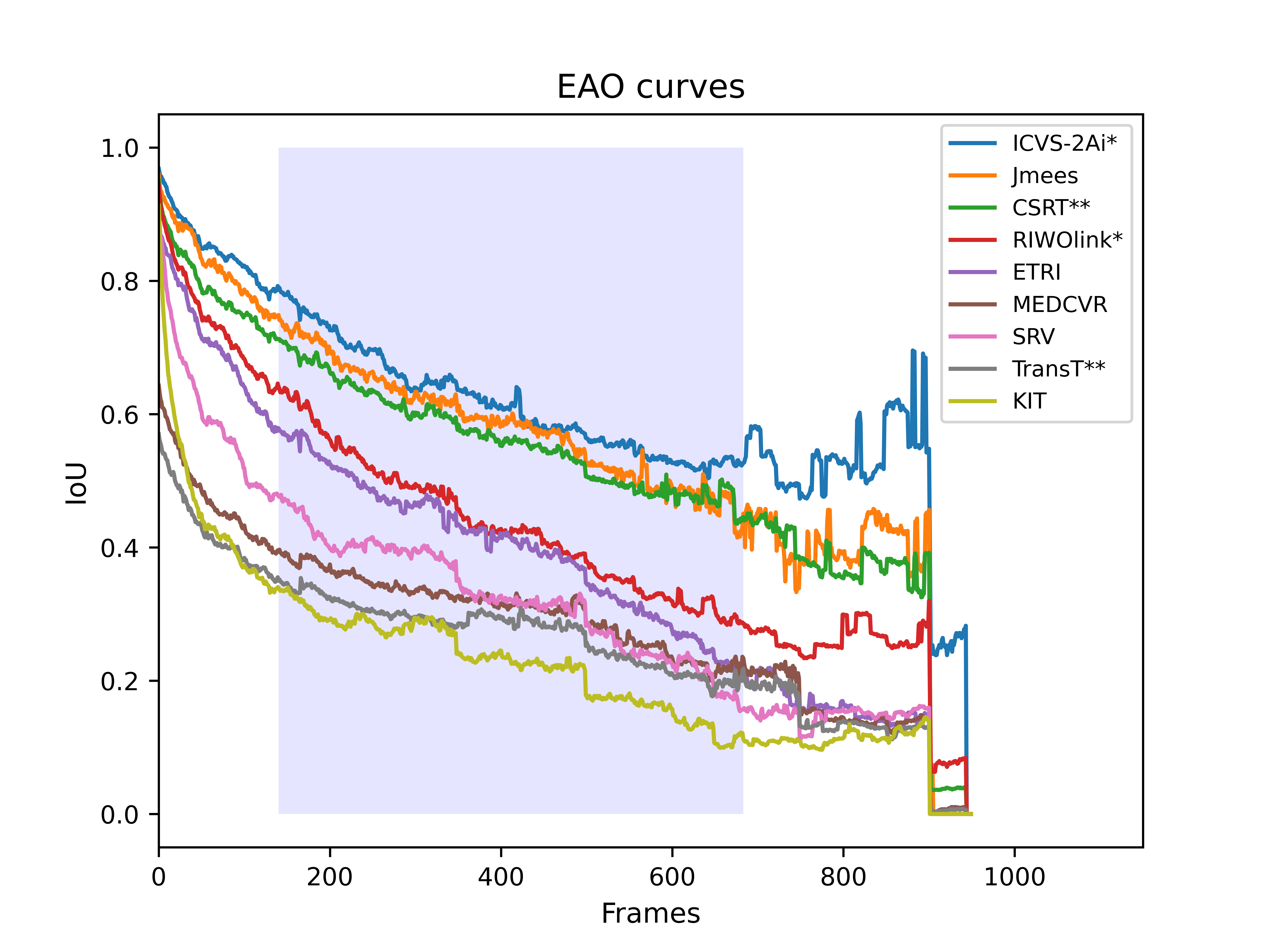}
\caption{The EAO curves for each participating team on the test subset. Highlighted in blue is the range of frames contributing to the EAO score, used for final ranking, from frames $N_{min} = 140$ to $N_{max} = 683$. Note that after frame 750, the IoU scores are noisier. This is because only a subset of the anchor sequences (10\%) are long enough to reach these higher frame numbers; therefore, these scores are averaged over a smaller number of frames. All cases reach at least 750 frames, but it is mostly case 3 - which has longer videos - that contributes to the last scores. ICVS-2AI had particularly good scores in case 3, hence the results increasing substantially at the end. Again, these scores are ignored, as otherwise it would negatively impact the evaluation reliability.}
\label{fig:eao_curve}
\end{figure*}

\begin{figure}[h]
\centering
\includegraphics[width=0.95\columnwidth]{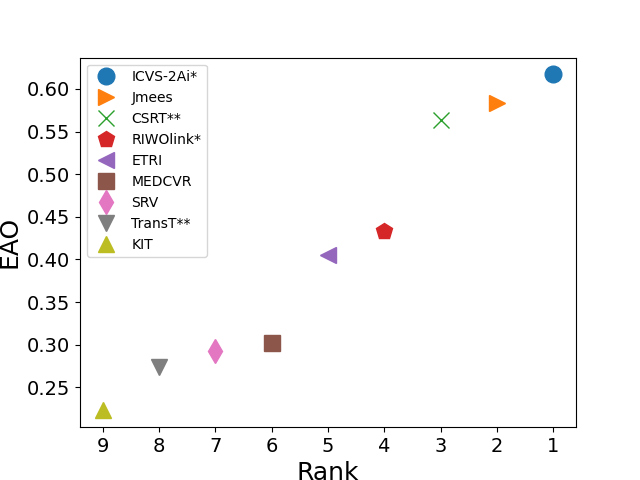}
\caption{EAO score vs. rank of each contestant on the test subset.}
\label{fig:rank_graph}
\end{figure}

\begin{figure}[h]
\centering
\includegraphics[width=0.95\columnwidth]{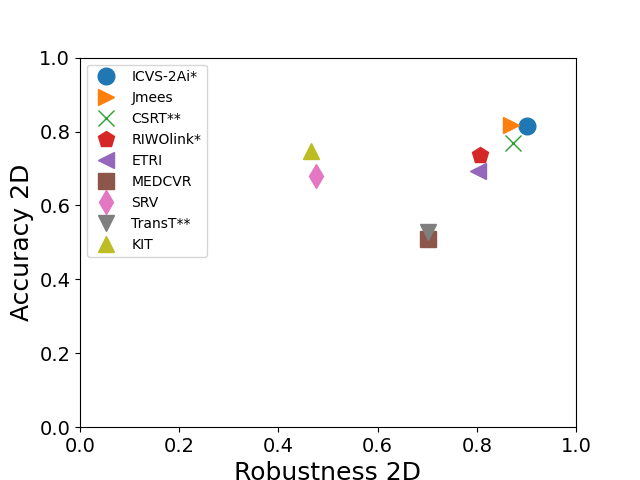}
\caption{The accuracy-robustness plot on the test data. The better the performance, the closer to the top right corner.}
\label{fig:ar_2d_plot}
\end{figure}

In the test subset (Table \ref{tab:test_res}), the EAO scores were highest for the ICVS-2Ai method with a score of 0.617. The Jmees method was next with a score of 0.583, and CSRT followed closely at 0.563. These scores indicate the superior overlap consistency of these methods during the testing phase. The associated EAO curves for the test subset are shown in Fig.~\ref{fig:eao_curve}. The final rankings can be found in Fig.~\ref{fig:rank_graph}, with the associated 2D accuracy vs. robustness shown in Fig.~\ref{fig:ar_2d_plot}.
The average scores on the validation subset are presented in Table~\ref{tab:val_res}.

%\add{As the top three methods produce similar EAO scores, to highlight a significant difference in their performances, we display in Fig??? the 3D error and robustness scores of the methods from test set Case 5. We deem this the most significant test of the methods as this is the only Case to use a handheld camera - producing a much more challenging tasks. For downstream tasks, the 3D error is blah}

%Note that in the tables and figures there are also post-challenge submissions scores, tagged with an $*$. In this challenge, post-challenge submissions were not considered for the awards. Therefore, \nth{1} place was awarded to `Jmees', which receive a GPU award from NVIDIA\textsuperscript{\tiny\textregistered}, \nth{2} place to ETRI, which receive a 1000\$ award from Intuitive\textsuperscript{\tiny\textregistered}, and \nth{3} place to MEDCVR, which received a 500\$ award from Intuitive\textsuperscript{\tiny\textregistered}.

\begin{sidewaystable*}
\centering
%\begin{table}[]
\caption{Case breakdown of tracking scores on the test subset.}
\begin{threeparttable}
\begin{tabular}{lccccccccccccccc}    
 & \multicolumn{15}{c}{\textbf{Test 2D Case Scores}}\\ \cline{2-16} 
\multicolumn{1}{c|}{}  & \multicolumn{3}{c|}{Case 1}& \multicolumn{3}{c|}{Case 2}& \multicolumn{3}{c|}{Case 3}& \multicolumn{3}{c|}{Case 4}& \multicolumn{3}{c|}{Case 5}\\ \cline{2-16} 
\multicolumn{1}{c|}{}& \multicolumn{1}{c|}{\multirow{2}{*}{\begin{tabular}[c]{@{}c@{}}\textit{Rob.}\\ \textit{2D}\end{tabular}}} & \multicolumn{1}{c|}{\multirow{2}{*}{\begin{tabular}[c]{@{}c@{}}Acc.\\ \textit{2D}\end{tabular}}} & \multicolumn{1}{c|}{\multirow{2}{*}{\begin{tabular}[c]{@{}c@{}}\textit{Error 2D}\\ {[}$pixels${]}\end{tabular}}} & \multicolumn{1}{c|}{\multirow{2}{*}{\begin{tabular}[c]{@{}c@{}}\textit{Rob.}\\ \textit{2D}\end{tabular}}} & \multicolumn{1}{c|}{\multirow{2}{*}{\begin{tabular}[c]{@{}c@{}}Acc.\\ \textit{2D}\end{tabular}}} & \multicolumn{1}{c|}{\multirow{2}{*}{\begin{tabular}[c]{@{}c@{}}\textit{Error 2D}\\ {[}$pixels${]}\end{tabular}}} & \multicolumn{1}{c|}{\multirow{2}{*}{\begin{tabular}[c]{@{}c@{}}\textit{Rob.}\\ \textit{2D}\end{tabular}}} & \multicolumn{1}{c|}{\multirow{2}{*}{\begin{tabular}[c]{@{}c@{}}Acc.\\ \textit{2D}\end{tabular}}} & \multicolumn{1}{c|}{\multirow{2}{*}{\begin{tabular}[c]{@{}c@{}}\textit{Error 2D}\\ {[}$pixels${]}\end{tabular}}} & \multicolumn{1}{c|}{\multirow{2}{*}{\begin{tabular}[c]{@{}c@{}}\textit{Rob.}\\ \textit{2D}\end{tabular}}} & \multicolumn{1}{c|}{\multirow{2}{*}{\begin{tabular}[c]{@{}c@{}}Acc.\\ \textit{2D}\end{tabular}}} & \multicolumn{1}{c|}{\multirow{2}{*}{\begin{tabular}[c]{@{}c@{}}\textit{Error 2D}\\ {[}$pixels${]}\end{tabular}}} & \multicolumn{1}{c|}{\multirow{2}{*}{\begin{tabular}[c]{@{}c@{}}\textit{Rob.}\\ \textit{2D}\end{tabular}}} & \multicolumn{1}{c|}{\multirow{2}{*}{\begin{tabular}[c]{@{}c@{}}Acc.\\ \textit{2D}\end{tabular}}} & \multicolumn{1}{c|}{\multirow{2}{*}{\begin{tabular}[c]{@{}c@{}}\textit{Error 2D}\\ {[}$pixels${]}\end{tabular}}} \\ \cline{1-1}
\multicolumn{1}{|l|}{\textbf{Method}} & \multicolumn{1}{c|}{}& \multicolumn{1}{c|}{}& \multicolumn{1}{c|}{}& \multicolumn{1}{c|}{}& \multicolumn{1}{c|}{}& \multicolumn{1}{c|}{}& \multicolumn{1}{c|}{}& \multicolumn{1}{c|}{}& \multicolumn{1}{c|}{}& \multicolumn{1}{c|}{}& \multicolumn{1}{c|}{}& \multicolumn{1}{c|}{}& \multicolumn{1}{c|}{}& \multicolumn{1}{c|}{}& \multicolumn{1}{c|}{}\\ \hline
\multicolumn{1}{|l|}{ICVS-2Ai*}& \multicolumn{1}{c|}{\cellcolor{blue!25}0.97} & \multicolumn{1}{c|}{0.75} & \multicolumn{1}{c|}{10±5} & \multicolumn{1}{c|}{\cellcolor{blue!25}0.91} & \multicolumn{1}{c|}{0.77} & \multicolumn{1}{c|}{10±6} & \multicolumn{1}{c|}{\cellcolor{blue!25}0.92} & \multicolumn{1}{c|}{0.87} & \multicolumn{1}{c|}{5±3} & \multicolumn{1}{c|}{\cellcolor{blue!25}1.0} & \multicolumn{1}{c|}{\cellcolor{blue!25}0.90} & \multicolumn{1}{c|}{\cellcolor{blue!25}3±1} & \multicolumn{1}{c|}{0.69} & \multicolumn{1}{c|}{\cellcolor{blue!25}0.80} & \multicolumn{1}{c|}{\cellcolor{blue!25}4±3} \\ \hline
\multicolumn{1}{|l|}{Jmees}& \multicolumn{1}{c|}{0.92} & \multicolumn{1}{c|}{\cellcolor{blue!25}0.76} & \multicolumn{1}{c|}{\cellcolor{blue!25}7±3} & \multicolumn{1}{c|}{0.89} & \multicolumn{1}{c|}{0.77} & \multicolumn{1}{c|}{\cellcolor{blue!25}7±3} & \multicolumn{1}{c|}{0.88} & \multicolumn{1}{c|}{\cellcolor{blue!25}0.88} & \multicolumn{1}{c|}{5±2} & \multicolumn{1}{c|}{0.94} & \multicolumn{1}{c|}{\cellcolor{blue!25}0.90} & \multicolumn{1}{c|}{\cellcolor{blue!25}3±1} & \multicolumn{1}{c|}{\cellcolor{blue!25}0.70} & \multicolumn{1}{c|}{0.75} & \multicolumn{1}{c|}{5±3} \\ \hline
\multicolumn{1}{|l|}{CSRT**}& \multicolumn{1}{c|}{0.92} & \multicolumn{1}{c|}{0.72} & \multicolumn{1}{c|}{\cellcolor{blue!25}7±3} & \multicolumn{1}{c|}{0.87} & \multicolumn{1}{c|}{0.77} & \multicolumn{1}{c|}{8±3} & \multicolumn{1}{c|}{0.87} & \multicolumn{1}{c|}{0.81} & \multicolumn{1}{c|}{5±2} & \multicolumn{1}{c|}{\cellcolor{blue!25}1.0} & \multicolumn{1}{c|}{0.85} & \multicolumn{1}{c|}{\cellcolor{blue!25}3±1} & \multicolumn{1}{c|}{\cellcolor{blue!25}0.70} & \multicolumn{1}{c|}{0.68} & \multicolumn{1}{c|}{6±3} \\ \hline
\multicolumn{1}{|l|}{RIWOlink*} & \multicolumn{1}{c|}{0.89} & \multicolumn{1}{c|}{0.66} & \multicolumn{1}{c|}{9±5}      & \multicolumn{1}{c|}{0.71} & \multicolumn{1}{c|}{\cellcolor{blue!25}0.78} & \multicolumn{1}{c|}{9±6}      & \multicolumn{1}{c|}{0.85} & \multicolumn{1}{c|}{0.86} & \multicolumn{1}{c|}{\cellcolor{blue!25}3±1}      & \multicolumn{1}{c|}{0.99} & \multicolumn{1}{c|}{0.69} & \multicolumn{1}{c|}{10±5}     & \multicolumn{1}{c|}{0.55} & \multicolumn{1}{c|}{0.59} & \multicolumn{1}{c|}{9±6}      \\ \hline
\multicolumn{1}{|l|}{ETRI}& \multicolumn{1}{c|}{0.88} & \multicolumn{1}{c|}{0.56} & \multicolumn{1}{c|}{21±13}    & \multicolumn{1}{c|}{0.82} & \multicolumn{1}{c|}{0.69} & \multicolumn{1}{c|}{14±10}    & \multicolumn{1}{c|}{0.85} & \multicolumn{1}{c|}{0.82} & \multicolumn{1}{c|}{7±3}      & \multicolumn{1}{c|}{0.94} & \multicolumn{1}{c|}{0.73} & \multicolumn{1}{c|}{7±4}      & \multicolumn{1}{c|}{0.49} & \multicolumn{1}{c|}{0.60} & \multicolumn{1}{c|}{10±6}     \\ \hline
\multicolumn{1}{|l|}{MEDCVR}& \multicolumn{1}{c|}{0.74} & \multicolumn{1}{c|}{0.54} & \multicolumn{1}{c|}{7±5}      & \multicolumn{1}{c|}{0.65} & \multicolumn{1}{c|}{0.48} & \multicolumn{1}{c|}{16±12}    & \multicolumn{1}{c|}{0.82} & \multicolumn{1}{c|}{0.68} & \multicolumn{1}{c|}{4±4}      & \multicolumn{1}{c|}{0.77} & \multicolumn{1}{c|}{0.61} & \multicolumn{1}{c|}{10±7}     & \multicolumn{1}{c|}{0.46} & \multicolumn{1}{c|}{0.57} & \multicolumn{1}{c|}{6±4}      \\ \hline
\multicolumn{1}{|l|}{SRV} & \multicolumn{1}{c|}{0.39} & \multicolumn{1}{c|}{0.49} & \multicolumn{1}{c|}{27±13}    & \multicolumn{1}{c|}{0.19} & \multicolumn{1}{c|}{0.55} & \multicolumn{1}{c|}{16±11}    & \multicolumn{1}{c|}{0.71} & \multicolumn{1}{c|}{0.80} & \multicolumn{1}{c|}{11±6}     & \multicolumn{1}{c|}{0.98} & \multicolumn{1}{c|}{0.69} & \multicolumn{1}{c|}{13±5}     & \multicolumn{1}{c|}{0.07} & \multicolumn{1}{c|}{0.56} & \multicolumn{1}{c|}{22±12}    \\ \hline
\multicolumn{1}{|l|}{TransT**} & \multicolumn{1}{c|}{0.72} & \multicolumn{1}{c|}{0.46} & \multicolumn{1}{c|}{18±4}    & \multicolumn{1}{c|}{0.64} & \multicolumn{1}{c|}{0.42} & \multicolumn{1}{c|}{24±10}    & \multicolumn{1}{c|}{0.82} & \multicolumn{1}{c|}{0.61} & \multicolumn{1}{c|}{16±4}      & \multicolumn{1}{c|}{0.79} & \multicolumn{1}{c|}{0.61} & \multicolumn{1}{c|}{11±7}     & \multicolumn{1}{c|}{0.47} & \multicolumn{1}{c|}{0.48} & \multicolumn{1}{c|}{11±4}    \\  \hline
\multicolumn{1}{|l|}{KIT} & \multicolumn{1}{c|}{0.26} & \multicolumn{1}{c|}{0.61} & \multicolumn{1}{c|}{19±13}    & \multicolumn{1}{c|}{0.17} & \multicolumn{1}{c|}{0.61} & \multicolumn{1}{c|}{23±17}    & \multicolumn{1}{c|}{0.81} & \multicolumn{1}{c|}{0.84} & \multicolumn{1}{c|}{9±7}      & \multicolumn{1}{c|}{0.84} & \multicolumn{1}{c|}{0.71} & \multicolumn{1}{c|}{11±6}     & \multicolumn{1}{c|}{0.16} & \multicolumn{1}{c|}{0.58} & \multicolumn{1}{c|}{20±10} \\ \hline
    & \multicolumn{1}{l}{}      & \multicolumn{1}{l}{}      & \multicolumn{1}{l}{}& \multicolumn{1}{l}{}      & \multicolumn{1}{l}{}      & \multicolumn{1}{l}{}& \multicolumn{1}{l}{}      & \multicolumn{1}{l}{}      & \multicolumn{1}{l}{}& \multicolumn{1}{l}{}      & \multicolumn{1}{l}{}      & \multicolumn{1}{l}{}& \multicolumn{1}{l}{}      & \multicolumn{1}{l}{}      & \multicolumn{1}{l}{}\\
& \multicolumn{15}{c}{\textbf{Test 3D Case Scores}}\\ \cline{2-16} 
\multicolumn{1}{c|}{}& \multicolumn{3}{c|}{Case 1}& \multicolumn{3}{c|}{Case 2}& \multicolumn{3}{c|}{Case 3}& \multicolumn{3}{c|}{Case 4}& \multicolumn{3}{c|}{Case 5}\\ \cline{2-16} 
\multicolumn{1}{c|}{}&  \multicolumn{1}{c|}{\multirow{2}{*}{\begin{tabular}[c]{@{}c@{}}\textit{Rob.}\\ \textit{3D}\end{tabular}}} & \multicolumn{2}{c|}{\multirow{2}{*}{\begin{tabular}[c]{@{}c@{}}\textit{Error 3D}\\ {[}$mm${]}\end{tabular}}}   & \multicolumn{1}{c|}{\multirow{2}{*}{\begin{tabular}[c]{@{}c@{}}\textit{Rob.}\\ \textit{3D}\end{tabular}}} & \multicolumn{2}{c|}{\multirow{2}{*}{\begin{tabular}[c]{@{}c@{}}\textit{Error 3D}\\ {[}$mm${]}\end{tabular}}}   & \multicolumn{1}{c|}{\multirow{2}{*}{\begin{tabular}[c]{@{}c@{}}\textit{Rob.}\\ \textit{3D}\end{tabular}}} & \multicolumn{2}{c|}{\multirow{2}{*}{\begin{tabular}[c]{@{}c@{}}\textit{Error 3D}\\ {[}$mm${]}\end{tabular}}}   & \multicolumn{1}{c|}{\multirow{2}{*}{\begin{tabular}[c]{@{}c@{}}\textit{Rob.}\\ \textit{3D}\end{tabular}}} & \multicolumn{2}{c|}{\multirow{2}{*}{\begin{tabular}[c]{@{}c@{}}\textit{Error 3D}\\ {[}$mm${]}\end{tabular}}}   & \multicolumn{1}{c|}{\multirow{2}{*}{\begin{tabular}[c]{@{}c@{}}\textit{Rob.}\\ \textit{3D}\end{tabular}}} & \multicolumn{2}{c|}{\multirow{2}{*}{\begin{tabular}[c]{@{}c@{}}\textit{Error 3D}\\ {[}$mm${]}\end{tabular}}}   \\ \cline{1-1}
\multicolumn{1}{|c|}{\textbf{Method}} & \multicolumn{1}{c|}{}     & \multicolumn{2}{c|}{}   & \multicolumn{1}{c|}{}     & \multicolumn{2}{c|}{}   & \multicolumn{1}{c|}{}     & \multicolumn{2}{c|}{}   & \multicolumn{1}{c|}{}     & \multicolumn{2}{c|}{}   & \multicolumn{1}{c|}{}& \multicolumn{2}{c|}{}   \\ \hline
\multicolumn{1}{|l|}{ICVS-2Ai*}& \multicolumn{1}{c|}{0.97}  & \multicolumn{2}{c|}{4±3}& \multicolumn{1}{c|}{\cellcolor{blue!25}0.91} & \multicolumn{2}{c|}{\cellcolor{blue!25}2±1}& \multicolumn{1}{c|}{0.93} & \multicolumn{2}{c|}{\cellcolor{blue!25}1±0}& \multicolumn{1}{c|}{\cellcolor{blue!25}1.0} & \multicolumn{2}{c|}{\cellcolor{blue!25}1±1}& \multicolumn{1}{c|}{0.76} & \multicolumn{2}{c|}{6±5}\\\hline
\multicolumn{1}{|l|}{Jmees}& \multicolumn{1}{c|}{\cellcolor{blue!25}1.0}  & \multicolumn{2}{c|}{\cellcolor{blue!25}2±1}& \multicolumn{1}{c|}{0.85} & \multicolumn{2}{c|}{\cellcolor{blue!25}2±1}& \multicolumn{1}{c|}{0.93} & \multicolumn{2}{c|}{1±1}& \multicolumn{1}{c|}{0.95} & \multicolumn{2}{c|}{\cellcolor{blue!25}1±1}& \multicolumn{1}{c|}{0.72} & \multicolumn{2}{c|}{\cellcolor{blue!25}5±4}\\ \hline
\multicolumn{1}{|l|}{CSRT**}& \multicolumn{1}{c|}{0.91}  & \multicolumn{2}{c|}{5±4}& \multicolumn{1}{c|}{0.83} & \multicolumn{2}{c|}{2±1}& \multicolumn{1}{c|}{0.93} & \multicolumn{2}{c|}{1±1}& \multicolumn{1}{c|}{\cellcolor{blue!25}1.0} & \multicolumn{2}{c|}{\cellcolor{blue!25}1±1}& \multicolumn{1}{c|}{\cellcolor{blue!25}0.78} & \multicolumn{2}{c|}{9±16}\\ \hline
\multicolumn{1}{|l|}{RIWOlink*} & \multicolumn{1}{c|}{0.96} & \multicolumn{2}{c|}{7±4}& \multicolumn{1}{c|}{0.81} & \multicolumn{2}{c|}{6±5}& \multicolumn{1}{c|}{0.93} & \multicolumn{2}{c|}{1±1}& \multicolumn{1}{c|}{0.99} & \multicolumn{2}{c|}{5±34}    & \multicolumn{1}{c|}{0.72} & \multicolumn{2}{c|}{11±7}    \\ \hline
\multicolumn{1}{|l|}{ETRI}& \multicolumn{1}{c|}{0.97} & \multicolumn{2}{c|}{6±4}& \multicolumn{1}{c|}{0.90} & \multicolumn{2}{c|}{7±6}& \multicolumn{1}{c|}{0.94} & \multicolumn{2}{c|}{2±3}& \multicolumn{1}{c|}{\cellcolor{blue!25}1.0}  & \multicolumn{2}{c|}{3±3}& \multicolumn{1}{c|}{0.71} & \multicolumn{2}{c|}{13±13}   \\ \hline
\multicolumn{1}{|l|}{MEDCVR}& \multicolumn{1}{c|}{0.90} & \multicolumn{2}{c|}{8±22}    & \multicolumn{1}{c|}{0.75} & \multicolumn{2}{c|}{20±125}  & \multicolumn{1}{c|}{0.88} & \multicolumn{2}{c|}{2±9}& \multicolumn{1}{c|}{0.92} & \multicolumn{2}{c|}{10±62}   & \multicolumn{1}{c|}{0.67} & \multicolumn{2}{c|}{14±24}   \\ \hline
\multicolumn{1}{|l|}{SRV} & \multicolumn{1}{c|}{0.66} & \multicolumn{2}{c|}{16±13}   & \multicolumn{1}{c|}{0.37} & \multicolumn{2}{c|}{16±12}   & \multicolumn{1}{c|}{0.96} & \multicolumn{2}{c|}{3±3}& \multicolumn{1}{c|}{\cellcolor{blue!25}1.0}  & \multicolumn{2}{c|}{2±1}& \multicolumn{1}{c|}{0.46} & \multicolumn{2}{c|}{18±66}   \\ \hline
\multicolumn{1}{|l|}{TransT**} & \multicolumn{1}{c|}{0.91} & \multicolumn{2}{c|}{23±8}   & \multicolumn{1}{c|}{0.82} & \multicolumn{2}{c|}{24±53}   & \multicolumn{1}{c|}{0.92} & \multicolumn{2}{c|}{6±14}& \multicolumn{1}{c|}{0.92}  & \multicolumn{2}{c|}{10±56}& \multicolumn{1}{c|}{0.70} & \multicolumn{2}{c|}{31±26}   \\ \hline
\multicolumn{1}{|l|}{KIT} & \multicolumn{1}{c|}{0.68} & \multicolumn{2}{c|}{20±16}   & \multicolumn{1}{c|}{0.70} & \multicolumn{2}{c|}{23±14}   & \multicolumn{1}{c|}{\cellcolor{blue!25}0.97} & \multicolumn{2}{c|}{2±2}& \multicolumn{1}{c|}{\cellcolor{blue!25}1.0}  & \multicolumn{2}{c|}{6±5}& \multicolumn{1}{c|}{0.68} & \multicolumn{2}{c|}{20±19}   \\ \hline
\end{tabular}
\begin{tablenotes}
    \item[$*$] post-challenge submission.
    \item[$**$] baseline method.
\end{tablenotes}
\end{threeparttable}
\label{tab:big_results}
%\end{table}
\end{sidewaystable*}

\begin{table*}[]
\centering
\caption{Challenge validation subset scores.}
\begin{threeparttable}
\begin{tabular}{lcccccc} & \multicolumn{5}{c}{\textbf{Validation}} & \multicolumn{1}{l}{} \\ \cline{2-6}
\multicolumn{1}{c|}{}  & \multicolumn{1}{c|}{\multirow{2}{*}{\textit{Rob. 2D}}} & \multicolumn{1}{c|}{\multirow{2}{*}{\textit{Acc. 2D}}} & \multicolumn{1}{c|}{\multirow{2}{*}{\begin{tabular}[c]{@{}c@{}}\textit{Error 2D}\\ {[}$pixels${]}\end{tabular}}} & \multicolumn{1}{c|}{\multirow{2}{*}{\textit{Rob. 3D}}} & \multicolumn{1}{c|}{\multirow{2}{*}{\begin{tabular}[c]{@{}c@{}}\textit{Error 3D}\\ {[}$mm${]}\end{tabular}}} \\ \cline{1-1} \cline{7-7} 
\multicolumn{1}{|l|}{\textbf{Method}} & \multicolumn{1}{c|}{}& \multicolumn{1}{c|}{}& \multicolumn{1}{c|}{}& \multicolumn{1}{c|}{}& \multicolumn{1}{c|}{} & \multicolumn{1}{c|}{\textbf{EAO}} \\ \hline
\multicolumn{1}{|l|}{Jmees}& \multicolumn{1}{c|}{\cellcolor{blue!25}0.724}& \multicolumn{1}{c|}{0.795}& \multicolumn{1}{c|}{5.5 ± 3.2} & \multicolumn{1}{c|}{0.741}& \multicolumn{1}{c|}{\cellcolor{blue!25}2.5 ± 2.0}& \multicolumn{1}{c|}{\cellcolor{blue!25}0.314}\\ \hline
\multicolumn{1}{|l|}{ICVS-2Ai*}& \multicolumn{1}{c|}{0.684}& \multicolumn{1}{c|}{\cellcolor{blue!25}0.830}& \multicolumn{1}{c|}{\cellcolor{blue!25}4.5 ± 2.3} & \multicolumn{1}{c|}{\cellcolor{blue!25}0.758}& \multicolumn{1}{c|}{6.9 ± 5.0}& \multicolumn{1}{c|}{0.295}\\ \hline
\multicolumn{1}{|l|}{CSRT**}& \multicolumn{1}{c|}{0.705}& \multicolumn{1}{c|}{0.766}& \multicolumn{1}{c|}{5.6 ± 3.1} & \multicolumn{1}{c|}{0.743}& \multicolumn{1}{c|}{3.9 ± 3.6}& \multicolumn{1}{c|}{0.292}\\ \hline
\multicolumn{1}{|l|}{ETRI}& \multicolumn{1}{c|}{0.635}& \multicolumn{1}{c|}{0.675} & \multicolumn{1}{c|}{14.1 ± 8.0} & \multicolumn{1}{c|}{0.720} & \multicolumn{1}{c|}{8.6 ± 9.3} & \multicolumn{1}{c|}{0.206}\\ \hline
\multicolumn{1}{|l|}{RIWOlink*}& \multicolumn{1}{c|}{0.665}& \multicolumn{1}{c|}{0.698}& \multicolumn{1}{c|}{9.5 ± 4.9} & \multicolumn{1}{c|}{0.749}& \multicolumn{1}{c|}{6.8 ± 6.0}& \multicolumn{1}{c|}{0.203}\\ \hline
\multicolumn{1}{|l|}{SRV}& \multicolumn{1}{c|}{0.397}& \multicolumn{1}{c|}{0.686}& \multicolumn{1}{c|}{11.7 ± 7.1} & \multicolumn{1}{c|}{0.721}& \multicolumn{1}{c|}{11.9 ± 12.7}& \multicolumn{1}{c|}{0.131}\\ \hline
\multicolumn{1}{|l|}{MEDCVR}& \multicolumn{1}{c|}{0.436}& \multicolumn{1}{c|}{0.614}& \multicolumn{1}{c|}{7.0 ± 5.6} & \multicolumn{1}{c|}{0.638}& \multicolumn{1}{c|}{18.4 ± 107.9}& \multicolumn{1}{c|}{0.110}\\ \hline
\multicolumn{1}{|l|}{KIT}& \multicolumn{1}{c|}{0.375}& \multicolumn{1}{c|}{0.672}& \multicolumn{1}{c|}{12.8 ± 8.0} & \multicolumn{1}{c|}{0.698}& \multicolumn{1}{c|}{17.2 ± 12.8}& \multicolumn{1}{c|}{0.103}\\ \hline
\multicolumn{1}{|l|}{TransT**}& \multicolumn{1}{c|}{0.410} & \multicolumn{1}{c|}{0.528} & \multicolumn{1}{c|}{13.1 ± 4.1} & \multicolumn{1}{c|}{0.673} & \multicolumn{1}{c|}{23.1 ± 25.8}& \multicolumn{1}{c|}{0.095}\\ \hline
\end{tabular}
\begin{tablenotes}
    \item[$*$] post-challenge submission; \textit{Rob.} robustness; \textit{Acc.} accuracy.
    \item[$**$] baseline method.
\end{tablenotes}
\end{threeparttable}
\label{tab:val_res}
\end{table*}

\begin{table*}[]
\centering
\caption{Challenge test subset scores.}
\begin{threeparttable}
\begin{tabular}{lcccccc}
 & \multicolumn{5}{c}{\textbf{Test}} & \multicolumn{1}{l}{} \\ \cline{2-6}
\multicolumn{1}{c|}{}  & \multicolumn{1}{c|}{\multirow{2}{*}{\textit{\textit{Rob. 2D}}}} & \multicolumn{1}{c|}{\multirow{2}{*}{\textit{Acc. 2D}}} & \multicolumn{1}{c|}{\multirow{2}{*}{\begin{tabular}[c]{@{}c@{}}\textit{Error 2D}\\ {[}$pixels${]}\end{tabular}}} & \multicolumn{1}{c|}{\multirow{2}{*}{\textit{Rob. 3D}}} & \multicolumn{1}{c|}{\multirow{2}{*}{\begin{tabular}[c]{@{}c@{}}\textit{Error 3D}\\ {[}$mm${]}\end{tabular}}}\\ \cline{1-1} \cline{7-7} 
\multicolumn{1}{|l|}{\textbf{Method}} & \multicolumn{1}{c|}{}& \multicolumn{1}{c|}{}& \multicolumn{1}{c|}{}& \multicolumn{1}{c|}{}& \multicolumn{1}{c|}{}  & \multicolumn{1}{c|}{\textbf{EAO}} \\ \hline
\multicolumn{1}{|l|}{ICVS-2Ai*}& \multicolumn{1}{c|}{\cellcolor{blue!25}0.901}& \multicolumn{1}{c|}{\cellcolor{blue!25}0.818}& \multicolumn{1}{c|}{6.7 ± 3.7} & \multicolumn{1}{c|}{\cellcolor{blue!25}0.917}& \multicolumn{1}{c|}{\cellcolor{blue!25}2.3 ± 1.7} & \multicolumn{1}{c|}{\cellcolor{blue!25}0.617}\\
\hline
\multicolumn{1}{|l|}{Jmees}& \multicolumn{1}{c|}{0.868}& \multicolumn{1}{c|}{\cellcolor{blue!25}0.818}& \multicolumn{1}{c|}{\cellcolor{blue!25}5.3 ± 2.4} & \multicolumn{1}{c|}{0.878}& \multicolumn{1}{c|}{2.7 ± 1.9} & \multicolumn{1}{c|}{0.583}\\ \hline
\multicolumn{1}{|l|}{CSRT**}& \multicolumn{1}{c|}{0.872}& \multicolumn{1}{c|}{0.769}& \multicolumn{1}{c|}{5.7 ± 2.6} & \multicolumn{1}{c|}{0.894}& \multicolumn{1}{c|}{3.3 ± 3.8} & \multicolumn{1}{c|}{0.563}\\ \hline
\multicolumn{1}{|l|}{RIWOlink*}& \multicolumn{1}{c|}{0.807}& \multicolumn{1}{c|}{0.737}& \multicolumn{1}{c|}{8.0 ± 4.5} & \multicolumn{1}{c|}{0.894}& \multicolumn{1}{c|}{5.8 ± 9.2}& \multicolumn{1}{c|}{0.433}\\ \hline
\multicolumn{1}{|l|}{ETRI}& \multicolumn{1}{c|}{0.802}& \multicolumn{1}{c|}{0.693}& \multicolumn{1}{c|}{12.1 ± 7.4}& \multicolumn{1}{c|}{0.909}& \multicolumn{1}{c|}{5.7 ± 5.2} & \multicolumn{1}{c|}{0.405}\\ \hline
\multicolumn{1}{|l|}{MEDCVR}& \multicolumn{1}{c|}{0.702}& \multicolumn{1}{c|}{0.509}& \multicolumn{1}{c|}{7.9 ± 5.8} & \multicolumn{1}{c|}{0.832}& \multicolumn{1}{c|}{9.2 ± 39.7} & \multicolumn{1}{c|}{0.302}\\ \hline
\multicolumn{1}{|l|}{SRV} & \multicolumn{1}{c|}{0.476}& \multicolumn{1}{c|}{0.681}& \multicolumn{1}{c|}{15.4 ± 7.5}& \multicolumn{1}{c|}{0.710}& \multicolumn{1}{c|}{8.5 ± 13.7}& \multicolumn{1}{c|}{0.293}\\ \hline
\multicolumn{1}{|l|}{TransT**} & \multicolumn{1}{c|}{0.701}& \multicolumn{1}{c|}{0.529}& \multicolumn{1}{c|}{16.3 ± 5.1}  & \multicolumn{1}{c|}{0.861}& \multicolumn{1}{c|}{17.3 ± 26.7} & \multicolumn{1}{c|}{0.274}\\ \hline
\multicolumn{1}{|l|}{KIT} & \multicolumn{1}{c|}{0.465}& \multicolumn{1}{c|}{0.747}& \multicolumn{1}{c|}{12.5 ± 8.0}  & \multicolumn{1}{c|}{0.810}& \multicolumn{1}{c|}{11.9 ± 9.5} & \multicolumn{1}{c|}{0.223}\\ \hline
\end{tabular}
\begin{tablenotes}
    \item[$*$] post-challenge submission; \textit{Rob.} robustness; \textit{Acc.} accuracy.
    \item[$**$] baseline method.
\end{tablenotes}
\end{threeparttable}
\label{tab:test_res}
\end{table*}

\section{Results Analysis} \label{statistics}

The test results highlight noticeable trends in performance among the evaluated methods. In the 2D scores — assessing Robustness, Accuracy, and Pixel Error — the standout performers are ICVS-2Ai, Jmees, and CSRT. All three produce high Robustness and Accuracy scores. Notably, while CSRT trails slightly in Accuracy compared to the other two methods, it matches them in Robustness score. This indicates that these top three methods remain robust under challenging conditions, even with minor shifts in accuracy.
Conversely, at the bottom of the tables, the SRV and KIT methods register suboptimal Robustness scores. This indicates fragility to challenging surgical scenarios.

In the 3D scores, ICVS-2Ai, Jmees, CSRT maintain their strong performance, with RIWOlink and ETRI also producing competitive Robustness. However, heightened Error rates in the MEDCVR, SRV, TransT and KIT methods raise concerns about their stability across diverse scenarios. While the top methods (ICVS-2Ai and Jmees) estimate disparity to update the size of the bounding box, the other methods failed to leverage the stereo information and hence have higher errors and standard deviations.

The EAO scores further corroborate these observations. The superior EAO scores of ICVS-2Ai, Jmees, and CSRT indicate a consistent overlap with ground truth over the duration of the videos. Conversely, the smaller EAO scores of RIWOlink, ETRI, MEDCVR, SRV, TransT and KIT imply degradation in tracking accuracy over time or in specific scenarios.

Besides the test subset, the results on the validation subset show similar patterns, with the top three methods (ICVS-2Ai, Jmees, and CSRT) standing out with better scores, lower errors and lower standard deviations.

%\begin{figure}
%\label{main:case4}
%\hspace{-1em}
%\begin{minipage}{.49\linewidth}
%\centering
%\subfloat[]{\includegraphics[scale=.24]{case_1_v1.png}}
%\end{minipage}%
%\hspace{-.1em}
%\begin{minipage}{.49\linewidth}
%\centering
%\subfloat[]{\includegraphics[scale=.24]{case_2_v1.png}}
%\end{minipage}%
%\par\medskip

%\hspace{-1em}
%\begin{minipage}{.49\linewidth}
%\centering
%\subfloat[]{\includegraphics[scale=.24]{case_3_v1.png}}
%\end{minipage}
%\hspace{-1em}
%\begin{minipage}{.49\linewidth}
%\centering
%\subfloat[]{\includegraphics[scale=.24]{case_4_v1.png}}
%\end{minipage}%
%\par\medskip

%\hspace{-1em}
%\begin{minipage}{.49\linewidth}
%\centering
%\subfloat[]{\includegraphics[scale=.24]{case_5_v1.png}}
%\end{minipage}%
%\caption{The motion of the left bounding box's centre for video 1 of: (a) case 1, (b) case 2, (c) case 3, (d) case 4, and (e) case 5. In (e), the bounding box's motion is spread out over the image plane, possibly creating a more difficult tracking task.}
%\label{fig:tracker_motion}
%\end{figure}

\section{Discussion}

\subsection{Camera Handling and Tracking Difficulties:}
The results from the test subset distinctly highlight the challenges posed by Case 5. Unlike other cases where the camera is mounted on a robotic arm, Case 5 was captured using a handheld camera. As seen in Table \ref{tab:test_dataset_description} (highlighted in red), this handheld camera introduced significant shakes due to hand tremors. These tremors contrast with the more controlled movements of a robotically-held camera, which retains its position until further adjustments. Consequently, the hand-induced instabilities contributed to the increased tracking difficulty and subsequent lower performance scores.

Interestingly, the validation subset showed overall lower scores compared to the test subset. Particularly, Case 2 of the validation subset emerged as especially challenging due to its tissue scanning procedure. Here, not only do standard surgical instruments have the potential to occlude the tracking region, but the ultrasound probe — essential for tissue scanning — frequently moves across the tracked region causing tracking drift. Another challenge in case 2 is the rapid camera distance adjustment, which often results in interlaced video artefacts, as noted by \citet{cui2022caveats}. Due to these challenges, a significant number of frames in Case 2 were labelled as ``difficult".

Surprisingly, Case 4 of the test subset — despite its low-resolution videos of a beating heart — did not pose as much difficulty as anticipated. Two factors facilitated this outcome. Firstly, the camera remained static throughout the videos, having been pre-adjusted for an optimal heart view. Secondly, the heart's rhythmic beats caused the same tissue region to reappear over identical image locations, making tracking more consistent. Overall, our observations confirm that increased camera movement — especially with handheld cameras — intensifies tracking challenges. This revelation suggests potential avenues for further research, such as determining which surgical procedures favour tracking. %or utilizing robot-provided data (e.g., camera kinematics) to enhance tracking and detect obstructions, whether from surgical instruments, ultrasound probes, or other devices.}

 %\delete{Regarding the performance evaluation results, it is noticeable that the overall scores on the validation subset (Tab.~\ref{tab:val_res}) are lower than the scores on the test subset (Tab.~\ref{tab:test_res}). The most evident difference between the two subsets is the average 3D velocity of the bounding box, which is greater in the validation subset (highlighted in blue in Tab.~\ref{tab:validation_dataset_description} and Tab.~\ref{tab:test_dataset_description}). Therefore, it is hypothesised that the velocity of the keypoint in the 3D space is related to the tracking performance. Nevertheless, despite the increased difficulty, the team ranking on the validation subset is similar to the ranking on the test subset. This indicates that there was no team that overfitted to the validation subset since the relative scores were appropriately translated to the test scores.}

%\delete{Concretely, on the test subset (Tab.~\ref{tab:big_results}), the most stand-out point to be noted is that case 5 seems to be the most challenging, registering the lowest performance for all contesting teams. One hypothesises is that this may be due to: (1) the large average 3D velocity of the bounding box (1.65 $mm/frame$, shown in Tab.~\ref{tab:test_dataset_description}), (2) the large motion of the bounding box over the image plane (shown in Fig.~\ref{fig:tracker_motion}), and (3) the low texture in the image region around the bounding box (shown in Fig.~\ref{fig:bbox_samples}).}

\subsection{Role of Deep Learning in Soft-Tissue Tracking}

From the overall results, it is clear that it is possible to develop tracking algorithms using unsupervised deep learning frameworks. However, there is clearly scope for improvement. Only ICVS-2Ai and Jmees were able to beat the baseline CSRT tracker, however, Jmees utilised the baseline CSRT as the main backbone of their algorithm, instead of deep learning. Therefore, only ICVS-2Ai's tracker was able to beat the CSRT tracker via unsupervised deep learning, emphasizing the difficulty of the task even when using state-of-the-art deep learning methods.
%\delete{Unlike other vision applications, for this challenge, the best submission (Jmees team) utilised the baseline CSRT as the main component of their algorithm, instead of deep learning. This goes against what is generally expected, as most computer vision topics have now been dominated by deep learning. We hope this finding stimulates interest in this problem for future research work. In fact, only ICVS-2Ai's tracker was able to beat the CSRT tracker via unsupervised learning.}

The other baseline tracker, TransT, was used to check whether it is possible to train a supervised-learning tracker using only non-surgical datasets, given that there are many available natural scene datasets with ground truth annotations. However, the low scores on both validation and test subsets indicate that TransT does not generalize well onto SurgT. This highlights the need for unsupervised methods for surgical data, given the lack of ground truth.

It is worth noting that all submissions track the left view's bounding box separately from the right view's bounding box. Therefore we believe that the stereo information was not fully exploited by the challenge participants. Given the vast research in multi-view geometry, there is an opportunity to better utilise stereo information. It would be interesting to explore whether a tracker that jointly estimates the bounding boxes on both left and right views would obtain better results. 

In this challenge, we observed that the traditional baseline method (CSRT) was still competitive against the more recent deep learning methods. This observation is somewhat counter-intuitive, considering the recent dominance of deep learning across various computer vision domains. As we anticipate future iterations of the SurgT challenge, it sparks curiosity about whether unsupervised learning methods will eventually achieve clearly distinctive top scores. There also exists the potential to combine deep learning with multi-view geometry, blending the best of both approaches.

\subsection{Future Prospects}

\textbf{3D Metrics Enhancement}:

The emerging applications in surgical tissue tracking predominantly demand precise 3D tracking capabilities. For instance, autonomous tissue scanning requires the surgical robot to accurately identify the 3D tissue position for precise imaging probe placement; autonomous suturing needs the exact positioning of the needle within the tissue; and the precision and usability of augmented reality overlays on soft tissue entirely depends on the accuracy of the tracking of the tissue's surface. Given this context, the challenge's authors assert that future iterations of SurgT should prioritise 3D metrics. The teams' rankings should thus pivot towards employing a 3D EAO score rather than its 2D counterpart.

A straightforward approach to computing a 3D accuracy score might involve utilizing a volumetric IoU instead of the conventional 2D bounding box-based method. In essence, this would require projecting the tracked keypoints from both stereo images into a 3D sphere and subsequently overlapping this projected sphere with the ground truth sphere (the one illustrated in Fig. \ref{fig:data_bbox}). It is important to note that metric development is critical to achieving translation success for computer vision in surgery \citep{maier2022metrics}, owing to the uniqueness of the task and data and the requirement for high precision and reliability.

In this first version of SurgT, we set a threshold for 3D tracking failure based on an error exceeding 10 cm. However, this threshold is considerably large for clinical applications such as robotic surgery. This elevated threshold also confuses the comparative analysis of various methods' 3D Robustness scores. For instance, in Case 4 of our test subset, many methods scored a 3D robustness of 1.0, blocking the differentiation of the methods. To better compare these methods, we propose the introduction of stricter thresholds in future SurgT benchmarks.

Another important consideration is that video resolution inversely affects the accuracy of 3D depth derived from disparity measurements. If the focus does shift to 3D metrics, it might require adapting the dataset to meet a minimum pixel resolution.

\textbf{Dataset Improvement}:

One significant area of future work is the improvement and expansion of the SurgT dataset, focusing on real in vivo human surgery. This includes not just adding more videos, but ensuring a wider range of surgical procedures and scenarios being covered. %For example, one question that remains open is how would it be possible to create ground-truth data for homogeneous tissue regions, since there is no distinct keypoint to label throughout the video.

\textbf{Prospects and Applications of Surgical Tissue Tracking}:

While there is ample opportunity for enhancement, our challenge results underscore the feasibility of tracking surgical tissue using unsupervised deep learning. This holds promise not just for aiding surgeons with augmented reality overlay but also in advancing surgical robotic systems — all thanks to the stereo endoscopes which are predominant in Minimally Invasive Surgery. The vast potential of our SurgT benchmark and dataset should catalyse further surgical technology research in tissue tracking — a fundamental challenge in surgical vision. We anticipate our contribution will motivate the community to adapt their techniques to this unique domain.

%Given the complexities and specific requirements of surgical procedures, computer vision techniques need to be especially refined and reliable. The success of tissue tracking in surgeries could have myriad applications, from assisting surgeons to improving the outcomes of surgeries. We encourage further exploration in this promising domain.

\section{Conclusions}  \label{conclusions}

This paper introduces the SurgT challenge, which is part of the Endoscopic Vision Challenge, organised in conjunction with MICCAI2022. First, the SurgT benchmark framework is introduced. A framework is proposed for evaluating and ranking the performance of tissue-tracking algorithms applied to surgical videos, for both monocular and stereo scenarios. This is expected to be used in the future as the standard benchmark framework for research and development of soft-tissue tracking algorithms. As this is a novel application, a dataset was curated for this work, focusing primarily on the annotation of keypoints for the validation of tracking methods. The data and code for benchmarking and labelling can be found at \url{https://surgt.grand-challenge.org/}. The aim of this challenge was to investigate unsupervised algorithms for soft-tissue tracking. The challenge attracted 5 teams, with 2 further post-challenge team submissions. Our study shows that in both 2D and 3D scenarios, methods such as ICVS-2Ai, Jmees, and CSRT consistently perform well. However, as camera movement increases and tracking scenarios become more complex, the challenges of accurate tracking become more evident. This emphasizes the need for a benchmark, such as SurgT, to support the development of tracking methods under demanding real-world surgery.

%\add{Based on the results from the inaugural year of this challenge, it is clear that traditional computer vision methods continue to maintain competitive performance in the field. The baseline used, a non-deep learning based method known as `CSRT', demonstrated robustness and accuracy that was on par with the top scoring methods. In fact, the method that secured the highest score actually incorporated `CSRT' as its backbone, only scoring marginally higher. These findings underscore the relevance and resilience of traditional computer vision methods even as we venture deeper into the era of unsupervised deep learning algorithms.}

\subsection{CRediT authorship contribution statement}
\textbf{Jo\~ao Cartucho}: Conceptualization, Methodology, Software, Resources, Data Curation, Formal analysis, Investigation, Validation, Visualization, Funding acquisition, Writing - Review \& Editing, Project administration.
\textbf{Alistair Weld}: Conceptualization, Methodology, Software, Data Curation, Formal analysis, Investigation, Validation, Visualization, Writing - Review \& Editing, Project administration.
\textbf{Samyakh Tukra}: Conceptualization, Data Curation, Writing - Original Draft.
\textbf{Haozheng Xu}: Software, Writing - Original Draft.
\textbf{Hiroki Matsuzaki}: 
Investigation, Writing - Original Draft.
\textbf{Taiyo Ishikawa}: 
Investigation, Writing - Original Draft.
\textbf{Minjun Kwon}: 
Investigation, Writing - Original Draft.
\textbf{Yong Eun Jang}: 
Investigation, Writing - Original Draft.
\textbf{Kwang-Ju Kim}: 
Investigation, Writing - Original Draft.
\textbf{Gwang Lee}: 
Investigation, Writing - Original Draft.
\textbf{Bizhe Bai}: 
Investigation, Writing - Original Draft.
\textbf{Lueder Kahrs}: 
Investigation, Writing - Original Draft.
\textbf{Lars Boecking}: 
Investigation, Writing - Original Draft.
\textbf{Simeon Allmendinger}: 
Investigation, Writing - Original Draft.
\textbf{Leopold M{\"u}ller}:  
Investigation, Writing - Original Draft.
\textbf{Yitong Zhang}: 
Investigation, Writing - Original Draft.
\textbf{Yueming Jin}: 
Investigation, Writing - Original Draft.
\textbf{Sophia Bano}: 
Investigation, Writing - Original Draft.
\textbf{Francisco Vasconcelos}: 
Investigation, Writing - Original Draft.
\textbf{Wolfgang Reiter}: 
Investigation, Writing - Original Draft.
\textbf{Jonas Hajek}: 
Investigation, Writing - Original Draft.
\textbf{Bruno Silva}: 
Investigation, Writing - Original Draft.
%\textbf{Lukas Buschle}: 
%Investigation, Writing - Original Draft.
\textbf{Estev\~{a}o Lima}: 
Investigation, Writing - Original Draft.
\textbf{Jo\~{a}o L. Vila\c{c}a}: 
Investigation, Writing - Original Draft.
\textbf{Sandro Queir\'os}: Software, Investigation, Writing - Review \& Editing.
\textbf{Stamatia Giannarou}: 
Supervision.

\subsection{Declaration of competing interest}
The authors declare that they have no known competing financial interests or personal relationships that could have appeared to influence the work reported in this paper.

\section*{Compliance with ethical standards}
\textbf{Conflict of interest} The authors declare that they have no conflict of interest to disclose. \\
\textbf{Ethical approval} All human and animal studies have been approved and performed in accordance with ethical standards.\\
\textbf{Informed consent} All the data used for this publication was previously publicly available online and was obtained with informed consent.\\
\textbf{Funding}
The authors are grateful for the sponsorship from Intuitive\textsuperscript{\tiny\textregistered} and NVIDIA\textsuperscript{\tiny\textregistered} for the prize awards.

\section{Acknowledgments}
The authors are grateful for the support from Intuitive\textsuperscript{\tiny\textregistered} and NVIDIA\textsuperscript{\tiny\textregistered} for sponsoring the prize awards. The authors are also grateful for the following research programs and grants: UK Research and Innovation (UKRI) Centre for Doctoral Training in AI for Healthcare (EP/S023283/1), the Royal Society (URF$\setminus$R$\setminus$201014), the Portuguese Foundation for Science and Technology (FCT; UIDB/50026/2020, UIDP/50026/2020, UIDB/05549/2020, UIDP/05549/2020, UIDB/50026/2020 and UIDP/50026/2020, and CEECIND/03064/2018), the Northern Portugal Regional Operational Programme (NORTE2020, under the Portugal 2020 Partnership Agreement, through the European Regional Development Fund; NORTE-01-0145-FEDER-000045 and NORTE-01-0145-FEDER- 000059), the company KARL STORZ SE \& Co. KG.
\bibliographystyle{model2-names.bst}\biboptions{authoryear}
\bibliography{refs}

\end{document}